\documentclass[10pt,journal,compsoc]{IEEEtran}
\ifCLASSOPTIONcompsoc
  \usepackage[nocompress]{cite}
\else
  \usepackage{cite}
\fi
\usepackage[colorlinks=true,linkcolor=black,citecolor=blue,urlcolor=blue,]{hyperref}
\usepackage[doipre={doi:~}]{uri}
\usepackage{booktabs}
\usepackage{makecell}
\usepackage{multirow}
\usepackage{cite}
\usepackage{xcolor}
\usepackage{caption}
\usepackage{subcaption}
\usepackage{colortbl}

\ifCLASSINFOpdf
   \usepackage[pdftex]{graphicx}
\else
\fi
\usepackage{amsmath}
\newcommand{\etal}{\textit{et al.}}

\newcommand{\eg}{\textit{e.g.}}
\usepackage{tikz}
\usepackage{tikz-qtree}
\usepackage{ragged2e}
\usepackage{lineno}

\begin{document}

\title{Deep Learning for Micro-expression Recognition:  A Survey}

\author{Yante~Li, Jinsheng~Wei, Yang~Liu,    Janne~Kauttonen~and~Guoying~Zhao*,~\IEEEmembership{Fellow, IEEE }% <-this % stops a space

\IEEEcompsocitemizethanks{
\IEEEcompsocthanksitem * represents the corresponding author.

\IEEEcompsocthanksitem Y. Li, J. Wei, Y. Liu, G. Zhao are with the Center for Machine Vision and Signal Analysis, University of Oulu, Oulu, FI-90014, Finland. E-mail:
{firstname.lastname}@oulu.fi
\IEEEcompsocthanksitem J. Wei is with the School of Telecommunications and Information Engineering, Nanjing University of Posts and Telecommunications, Nanjing 210003 China; Email: 2018010217@njupt.edu.cn.
\IEEEcompsocthanksitem J.~Kauttonen is with the School of Digital Business, Haaga-Helia   University of Applied Sciences, Helsinki, FI-00520, Finland; Email: Janne.Kauttonen@haaga-helia.fi.
}}

%\protect\\
%E-mail: see http://www.michaelshell.org/contact.html
%\IEEEcompsocthanksitem J. Doe and J. Doe are with Anonymous University.}% <-this % stops an unwanted space


\markboth{Journal of \LaTeX\ Class Files,~Vol.~14, No.~8, August~2015}%
{Shell \MakeLowercase{\textit{et al.}}: Bare Demo of IEEEtran.cls for Computer Society Journals}



\IEEEtitleabstractindextext{%
\begin{abstract}
Micro-expressions  (MEs)  are  involuntary  facial movements  revealing  people’s  hidden  feelings  in  high-stake  situations  and have practical importance in various fields. Early methods for Micro-expression Recognition (MER) are mainly based on traditional features. Recently, with the success of Deep Learning (DL) in various tasks, neural networks have received increasing interest in MER. Different from macro-expressions, MEs are spontaneous, subtle, and rapid facial movements, leading to difficult data collection and annotation, thus publicly available datasets are usually small-scale. Currently, various DL approaches have been proposed to solve the ME issues and improve MER performance. In this survey, we provide a comprehensive review of deep MER and define a new taxonomy for the field encompassing all aspects of MER based on DL, including datasets, each step of the deep MER pipeline,  and performance comparisons of the most influential methods. The basic approaches and advanced developments are summarized and discussed for each aspect. Additionally, we conclude the remaining challenges and potential directions for the design of robust MER systems.
Finally, ethical considerations in MER are discussed. To the best of our knowledge, this is the first survey of deep MER methods, and this survey can serve as a reference point for future MER research.
\end{abstract}

}

\maketitle

\IEEEdisplaynontitleabstractindextext

% \fi
\IEEEpeerreviewmaketitle

\IEEEraisesectionheading{\section{Introduction}\label{sec:introduction}}

\IEEEPARstart{F}{acial} 
expression (FE) is one of the most powerful and universal means for human communication, which is highly associated with human mental states, attitudes, and intentions.
 Besides ordinary FEs (also known as macro-expressions) that we see daily, emotions can also be expressed in a special format of Micro-expressions (MEs) under certain conditions. MEs are FEs revealing people’s  hidden  feelings in high-stake situations when people try to conceal their true feelings \cite{ekman2003darwin}. Different from macro-expressions, MEs are spontaneous, subtle, and rapid (1/25 to 1/3 second) facial movements reacting  to  emotional  stimulus \cite{Ekman:1971,Ekman:2009}. 
%thus promoting cultural and social inclusion among individuals coming from different realities and belonging to different categories, including disadvantaged and at-risk groups, as well as vulnerable people. 

\begin{figure}
  \centering
%  \centerline{\includegraphics[width=8.5cm]{flowImg.pdf}}
\includegraphics[width=\columnwidth]{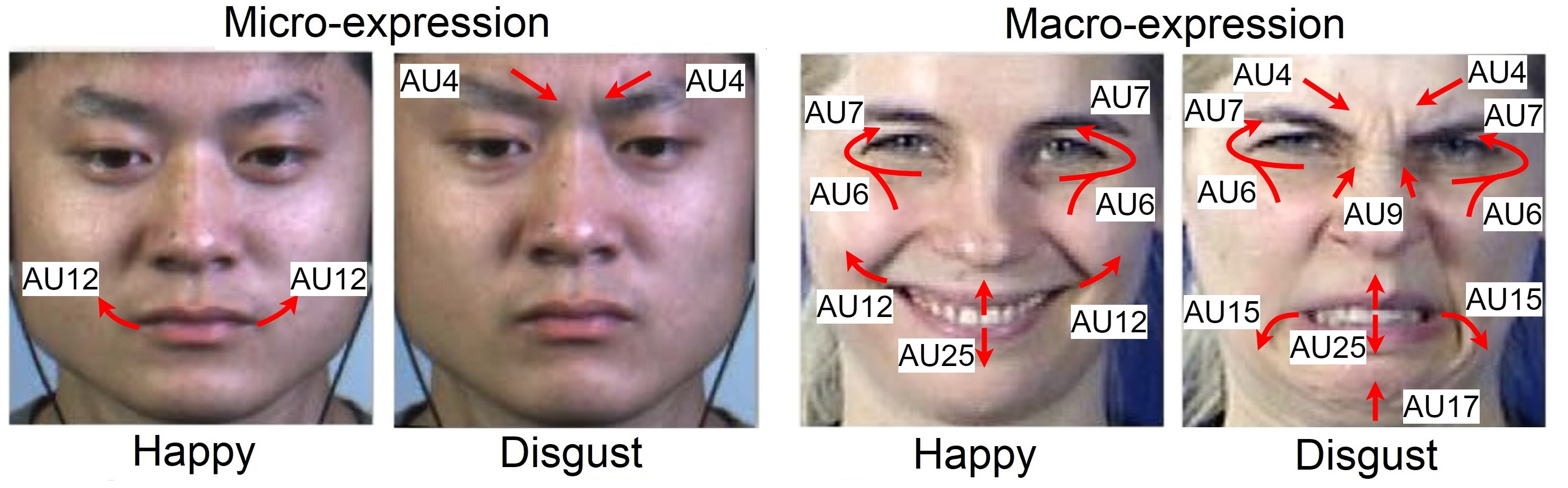}
\vspace{-4ex}
\caption{Examples of micro-expressions in CASME II \cite{Yan2014CASME} and macro-expressions in MMI \cite{pantic2005web}, as well as the active AUs. The red arrow represents the muscle movement direction. AU4, AU6, AU7, AU9, AU12, AU15, and AU25 represent brow lowerer, cheek raise, lids tight, nose wrinkle, lip corner puller,  lip corner depressor, and lips part, respectively. }
\vspace{-3ex}
\label{fig:MicroMacro}
\end{figure}

The ME phenomenon was firstly discovered by Haggard and Isaacs \cite{haggard1966micromomentar} in 1966. Three years later, Ekman and Friesen also declared the finding of MEs \cite{ekman1969nonverbal} during examining psychiatric patient's videos for lie detection.
In the following years, Ekman~\etal~ continued ME research and developed the Facial Action Coding System (FACS) \cite{friesen1978facial} and Micro Expression Training Tool (METT) \cite{ekman2002METT}. Specifically,  FACS breaks down FEs into individual components of muscle movement, called Action Units (AUs) \cite{friesen1978facial}.
AU analysis can effectively resolve the ambiguity issue to represent individual expression and increase Facial Expression Recognition (FER) performance \cite{xuesongCVPR2019}. 
Fig.~\ref{fig:MicroMacro} shows the example of micro- and macro-expressions as well as activated AUs in each FE. On the other hand, METT is helpful for increasing people's emotional awareness. \textcolor{black}{It can promote manual ME detection performance which provides a potential chance  to build reliable ME datasets.}  
%\cite{wu2011machine}

%Later many other researchers joined the ME analysis.

%Different from macro-expressions,

\textcolor{black}{
MER is the task of classifying ME clips into various emotion categories. In each ME clip, the frame starting facial movements is denoted as the onset frame, while the end frame is the offset frame. The frame with the largest intensity is the apex frame. Like FER, MER also classifies facial images/sequences into categories such as anger, surprise, and happiness. However, MER is more challenging as spontaneous MEs are involuntary, subtle, and fleeting. In addition, MEs can also be impacted by emotional context and cultural background \cite{merghani2018review,crivelli2019inside,niedenthal2019historical}. Therefore, it is difficult to collect and annotate ME data, leading to small-scale ME datasets and existing methods are incapable of dealing with subtleness and fleetness. }

 MER has drawn increasing interest recently due to its practical importance in  many human-computer interaction systems. The first spontaneous MER research can be traced to Pfister~\etal's work~\cite{Pfister2011iccv} which  utilized a Local Binary Pattern from Three Orthogonal Planes (LBP-TOP)~\cite{Li2013A} on the first public spontaneous ME dataset: SMIC \cite{li2013spontaneous}.
Following the work of~\cite{Li2013A}, various approaches based on appearance and geometry features ~\cite{Li2015Towards,wei2021comparative}  were proposed for improving the performance of MER.

%and achieved very promising results that compare favorably with the human accuracy.

In recent years, with the advance of Deep Learning (DL) and its successful extensions on object detection \cite{liu2020deep}, human tracking \cite{brunetti2018computer,liu2018visual}, image retrieval  \cite{li2016fast,li2015more} and FER \cite{li2020deep,liu2021sg,liu2022uncertain}, researchers have started to exploit MER with DL. Although MER with DL becomes challenging because of the limited ME samples and low intensity, great progress on MER has been made through  designing effective shallow networks, exploring Generative Adversarial Net (GAN)~\cite{yu2020ice} and so on. Currently, DL-based MER has achieved the state-of-the-art performance. 
%received increasing attention and 

%Study (Ekman, 2002) shows that for micro-expression recognition tasks, ordinary people without training only perform slightly better than chance on average. So computer vision and machine learning methods for automatic micro-expression analysis become appealing. Pfister~\etal (2011) started pioneering research on spontaneous micro-expression recognition with the first publically available spontaneous micro-expression dataset: SMIC, and achieved very promising results that compare favorably with the human accuracy. Since then micro-expression study in computer vision field has been attracting attentions from more and more researchers. A number of works have been contributing to the automatic micro-expression analysis from the aspects of new datasets collection (from emotion level annotation to action unit level annotation; Li~\etal, 2013; Davison~\etal, 2018), micro-expression recognition (from signal apex fraMER to whole video recognition; Wang~\etal, 2015; Liu~\etal, 2016; Li Y.~\etal, 2018; Huang~\etal, 2019) and micro-expression detection (from micro-expression peak detection to micro-expression onset and offset detection; Patel~\etal, 2015; Xia~\etal, 2016; Jain~\etal, 2018). 

{\textcolor{black}{In this survey, we review the research on MER by DL since 2016 when the DL technology was firstly adopted in MER. Due to the page limitation, the representative works published in well-known journals and conferences, such as IEEE TPAMI, IEEE TAC, IEEE TIP, and ACM MM are specifically discussed. The ordinary FER approaches and MER with traditional learning methods are not considered in this survey.} Although a few MER surveys have discussed the historical evolution and algorithmic pipelines for MER \cite{xie2020overview,goh2020micro,takalkar2018survey,oh2018survey,ben2021video,zhou2021survey}, they mainly focus on traditional methods and only introduce some recent DL approaches. The DL-based MER has not been discussed systematically and specifically. As far as we know, this is the first survey of the DL-based MER. Different from previous surveys, we analyze the strengths and shortcomings of dynamic network inputs which are important for MER based on DL. Furthermore, the network blocks, architectures, training strategies, and losses are discussed and summarized in detail and future research directions are identified. The goal of this survey is to provide a DL-based MER dictionary that can serve as a reference point for future MER research.

\begin{figure}
  \centering
%  \centerline{\includegraphics[width=8.5cm]{flowImg.pdf}}
\includegraphics[width=8.7cm]{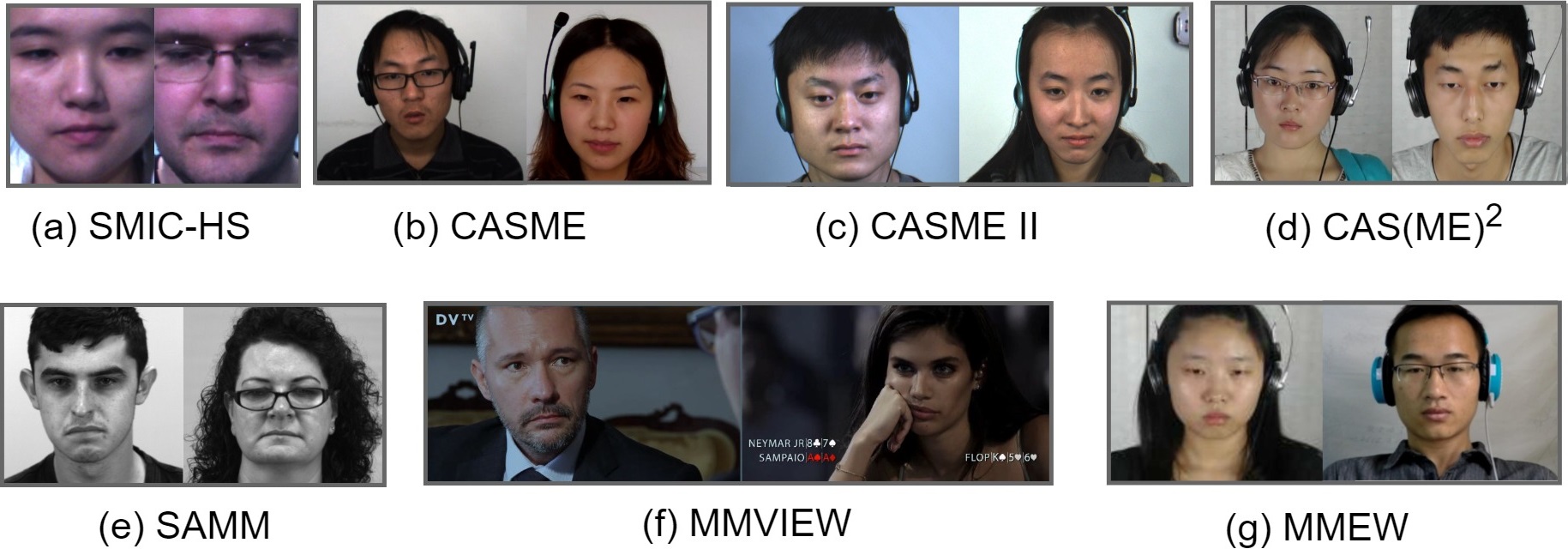}
\vspace{-1ex}
\caption{Examples of ME samples in ME datasets for MER.}
\vspace{-4ex}
\label{fig:microdatasets}
\end{figure}

\textcolor{black}{This paper is organized as follows: Section \ref{sec:Datasets} introduces spontaneous ME datasets. Section \ref{sec:taxonomy} presents the taxonomy we defined for MER based on DL. Section \ref{sec:inputsall} discusses the various inputs for deep MER.  Section \ref{sec:network} provides a detailed review of neural networks for MER.} 
The evaluation matrix, protocol, and the performance of representative DL-based MER are described 
in Section \ref{sec:experiments}.
Section \ref{sec:Challenge} summarizes current challenges and potential study directions.
Finally,  Section 8 discusses the ethical considerations.

\begin{table*}[t]
\renewcommand{\arraystretch}{1.3}
\centering
\small
\caption{Spontaneous Datasets for MER}
\label{dataset}
\scalebox{0.72}{ 
\begin{tabular}{|c|c|c|c|c|c|c|c|c|c|c|c|c|}  \bottomrule[1.5pt]
 Database    &Resolution&Facial size& Frame rate& Samples & subjects & Expression &AU&Apex&Eth&Env \\\bottomrule[1.5pt] 
%Canal9 \cite{vinciarelli2009canal9}& \\\hline
%Polikovsky \cite{polikovsky2009facial}&$640\times480$& N/A& 200&42&10&P&6&N&3&L\\\hline
%USF-HD \cite{shreve2011macro}& $720\times1280$&N/A &30 &100& N/A&P&3&N&N/A&L   \\\hline
%York-DDT \cite{warren2009detecting}&     \\\hline
\makecell[c]{SMIC \\ HS/NIR/VIS \cite{li2013spontaneous}} & $640\times480$ & $190\times230$& 100/25/25& 164/71/71 &16/8/8& \makecell[c]{ \textit{Pos} (51) \textit{Neg} (70) \textit{Sur} (43) / \textit{Pos} (28) \textit{Neg} (23) \textit{Sur} (20) /   \\  \textit{Pos} (28) \textit{Neg} (24) \textit{Sur} (19)} &$\circ$&$\circ$&3&\textit{L}\\ 
\hline

CASME \cite{Yan2013CASME}  & \makecell[c]{$640\times480$\\ $1280\times720$}   & $150\times90$ &60&195&35&\makecell[c]{ \textit{Hap} (5) \textit{Dis} (88) \textit{Sad} (6) \textit{Con} (3) \textit{Fea} (2) \\ \textit{Ten} (28) \textit{Sur} (20) \textit{Rep} (40)  }&$\bullet$&$\bullet$&1&\textit{L}\\\hline  

CASME II \cite{Yan2014CASME}& $640\times480$ &$250\times340$&200&247&35&\ \textit{Hap} (33) \textit{Sur} (25) \textit{Dis} (60)  \textit{Rep} (27)  \textit{Oth} (102)&$\bullet$&$\bullet$&1&\textit{L} \\\hline

CAS(ME)$^2$ \cite{qu2017cas}& $640\times480$& -&30& \makecell[c]{Macro 300  \\ Micro 57 }  &22&\textit{Hap} (51) \textit{Neg} (70) \textit{Sur} (43) \textit{Oth} (19)&$\bullet$&$\bullet$&1&\textit{L}   \\\hline 

SAMM~\cite{davison2018samm}&$2040\times1088$ & $400\times400$ &200 &159&32&\textit{Hap} (24) \textit{Ang} (20) \textit{Sur} (13) \textit{Dis} (8) \textit{Fea} (7) \textit{Sad} (3) \textit{Oth} (84)&$\bullet$&$\bullet$&13&\textit{L}  \\\hline

MEVIEW \cite{artnerspotting}& $720\times1280$ &- &25&31&16 & \textit{Hap} (6) \textit{Ang} (2) \textit{Sur} (9) \textit{Dis} (1) \textit{Fea} (3) \textit{Unc} (13) Con(6)  &$\bullet$&$\circ$&-&\textit{W}\\\hline 

MMEW~\cite{ben2021video}& $1920\times1080$&$400\times400$&90&300&36 & \textit{Hap} (36) \textit{Ang} (8) \textit{Sur} (80) \textit{Dis} (72) \textit{Fea} (16) \textit{Sad} (13) \textit{Oth} (102)  &$\bullet$ &$\bullet$&1&\textit{L}\\\hline

Composite ME~\cite{see2019megc}& \makecell[c]{$640\times480$\\ $1280\times720$ \\ $720\times1280$} &\makecell[c]{$150\times90$\\$250\times340$ \\ $400\times400$}  &200&442&68&\textit{Pos} (109), \textit{Neg} (250), and \textit{Sur} (83)&$\circ \bullet$&$\circ \bullet$&13&\textit{L} \\\hline

Compound ME~\cite{zhao2019convolutional}&\makecell[c]{$640\times480$\\ $1280\times720$ \\ $720\times1280$} &$150\times90$  &200&1050&90& \makecell[c]{ \textit{Neg} (233) \textit{Pos} (82) \textit{Sur} (70) \\ \textit{PS} (74) \textit{N S} (236) \textit{PN} (197) \textit{NN} (158)}&$\circ \bullet$&$\circ \bullet$&13&\textit{L} \\ \bottomrule[1.5pt]

%Synthetic ME& \\\hline

\end{tabular}}
{\raggedright $^1$   Eth: Ethnicity;  Env : Environment. \par
$^2$ \textit{Pos}: Positive; \textit{Neg}: Negative; \textit{Sur}: Surprise; \textit{Hap}: Happiness; \textit{Dis}: Disgust;  \textit{Rep}: Repression;  
\textit{Ang}: Anger;  \textit{Fea}: Fear; \textit{Sad}: Sadness; \textit{Con}: Contempt;
\textit{Unc}: Unclear; 
\textit{Oth}: Others; 
\textit{PS}: Positively surprise; \textit{NS} Negatively surprise;  \textit{PN}: Positively negative;  \textit{NN}: Negatively negative;
\textit{L}:Laboratory; \textit{W}:In the wild. \par
$^3$ $\circ$ represents unlabeled; $\bullet$  represents  labeled and - represents unknown  \par
 
 }
\end{table*}

%\textcolor{black}{
%\section{Preliminaries of MER with DL}
%MER is the task of classifying ME clips into various emotion categories.  In each ME clip, the frame starting facial movements is denoted as the onset frame, while the end frame is offset frame. The frame with the largest intensity is the apex frame.
%Similar to FER, they both classify facial images/sequences into  categories such as anger, surprise, happiness and so on.
%However, compared to FER, MER is a more challenging task as MEs are more subtle, rapid, and difficult to be collected. Since MEs are spontaneous facial movements always occurred under high-stake situations, it is hard to induce MEs leading to small-scale ME datasets. Thus, pre-processing steps, such as data augmentation and motion magnification, are important for MER. On the other hand, in terms of network design of DL, various efficient blocks, architectures, training strategies and losses for effective MER are presented. In the following sections, the key points for effective MER will be discussed specifically.  }

\section{Datasets}
\label{sec:Datasets}
Different from macro-expressions which can be easily captured in our daily life, 
MEs are involuntary brief FEs, particularly occurring under high stake situations. Four early databases appeared continuously around 2010: Canal9~\cite{vinciarelli2009canal9}, York-DDT \cite{polikovsky2009facial}, Polikvsky’s database \cite{polikovsky2009facial} and USF-HD \cite{shreve2011macro}. However, Canal9 and York-DDT are not aimed for ME research. Polikvsky’s database  and USF-HD include only posed MEs which are collected by asking participants to intentionally pose or
mimic a micro movement. The posed expressions contradict with the spontaneous nature of MEs.  Currently, these databases are not used anymore for MER.  In the recent years, several spontaneous
ME databases were created, including: SMIC~\cite{li2013spontaneous} and its extended version SMIC-E, CASME \cite{Yan2013CASME}, CASME II \cite{Yan2014CASME}, CAS(ME)$^2$ \cite{qu2017cas}, SAMM~\cite{davison2018samm}, and micro-and-macro expression
warehouse (MMEW)~\cite{ben2021video}. In this survey, we focus on the spontaneous datasets.

\textcolor{black}{
In a general ME dataset collection procedure, participants are asked to keep a poker face while watching video clips to induce spontaneous MEs. The video clips are selected according to previous psychological studies, which can elicit strong emotions. Commonly, a high-speed camera is utilized to record facial videos. After one participant watched a video clip, he/she fills in a self-report questionnaire to report his/her true feelings about the video clip. As well, considering cultural backgrounds may have an impact on MEs \cite{zhang2018emotional}, participants from different ethnicities could be recruited \cite{davison2018samm} for the potential study of cultural impact on MEs.}

\textcolor{black}{
Since the MEs are subtle and rapid, annotators are usually trained with FACS and certified facial action unit coders are employed to detect the MEs in the facial videos. The FACS helps people look precisely at the facial movements to make ME detection reliable. Specifically, when the duration of the facial action unit is less than 0.5s, the clip is regarded as a ME clip. The MEs are annotated into discrete categories. In SMIC \cite{li2013spontaneous}, the emotions are labeled as ‘positive’, ‘negative’, and ‘surprise’ according to the participants’ self-reports. However, mixed emotions may be induced while the participants watch one video clip. Annotations based on the general emotion reported after watching the video, which usually allows one emotion, are not accurate. To this end, several datasets, such as CASME \cite{Yan2013CASME} and CASME II \cite{Yan2014CASME}, consider AUs, self-reports, and the watched video clips to label the MEs. When there are ambiguities and conflicts in the emotion annotation, the emotion is annotated as ‘others’. Furthermore, to alleviate annotation bias caused by an individual annotator, the ME annotations are always carried out through cross-validation by multiple annotators.  }
The specific details of datasets are introduced as followings:

\textbf{SMIC}  \cite{li2013spontaneous} is consisted of three subsets: SMIC-HS, SMIC-VIS and SMIC-NIR.  SMIC-VIS and SMIC-NIR contain 71 samples recorded by normal speed cameras with 25 fps of visual (VIS) and near-inferred  light range (NIR), respectively. 
%SMIC-HS recorded by 100 fps high-speed cameras is used for MER.  SMIC-HS contains 164 spontaneous ME samples from 16 subjects. These samples are annotated  into three categories based on self-reports: \textit{positive}, \textit{negative}, and \textit{surprise}. %Moreover, the start and end frame of ME clips, denoted as onset and offset, are labeled. 

\textbf{CASME} \cite{Yan2013CASME} contains spontaneous 159 ME clips from 19 subjects including frames from onset to offset.   The emotions were labeled partly based on AUs and also taking account of  participants’ self-reports and the content of the video episodes. Besides the onset and offset, the apex frames are also labeled. The shortcoming of CASME is the imbalanced sample distribution among classes. 
%It is recorded by high-speed camera at 60 fps. 

%:  The collected data were analyzed with FACS and the unemotional facial movements were removed.\textit{happiness} (5 samples),  \textit{disgust} (88 samples), \textit{sadness} (6 samples), \textit{surprise} (20 samples), \textit{fear} (2 samples), \textit{tenseness}, \textit{repression} (40 samples), and \textit{contempt} (3 samples)

%Similar to SMIC, participants were requested to keep neutral when watching the simulating videos. Otherwise, the amount of token will be deducted. The shortcoming of CASME is  the imbalanced sample distribution among classes. 
%The duration of the displayed videos was about 1–4 min, and 

\textbf{CASME II}  \cite{Yan2014CASME} is an improved version of the CASME dataset. Samples in CASME II are increased to 247 MEs from 26  subjects and they are recorded by high-speed camera at 200 fps with face sizes cropped to $280 \times 340$. Thus, it has a greater temporal and spatial resolution, compared with CASME. 

%However, CASME II suffers from class imbalance and the subjects are limited to youths with the same ethnicity. 

%There are five kinds of ME expressions: \textit{happiness} (33 samples), \textit{surprise} (25 samples), \textit{disgust} (60 samples), \textit{repression} (27 samples), and \textit{others} (102 samples). However, CASME II suffers from class imbalance and the subjects are limited to youths with the same ethnicity. 

%Figure 1 shows one of the frame sequences of disgust in CASME II. Intensive selection was done to pick the best samples out of 2500 FEs.Afewvideos with a maximum duration of 3 minwere displayed to trigger participants’ emotion. In addition, recording is done at 200 frames per second to collect FE in a controlled environment.After recording, participants were required to report their feelings for tailoring purposes.  CASME II also resolved illumination problems present in the previous dataset by providing a steady and high-intensity lighting environment.

\textbf{CAS(ME)$^2$} \cite{qu2017cas} consists of spontaneous macro- and micro-expressions elicited from 22 subjects. CAS(ME)$^2$ has samples with longer durations which makes it suitable for ME spotting.  Compared to the above datasets, the samples in CAS(ME)$^2$ were recorded with a relatively low frame rate in a relatively small number of ME samples, which makes it unsuitable for DL approaches.

%The dataset is divided into two parts. Part A includes 87 long videos with both  macro- and micro-expressions and part B contains 300 cropped spontaneous macro-expression samples and 57 micro-expression samples.  Compared to above datasets, the samples in CAS(ME)$^2$ were recorded with a relatively low frame rate in relatively small number of ME samples, which makes it being unsuitable for DL approaches.

\textbf{SAMM} \cite{davison2018samm} collects 159 ME samples from 32 participants.  The samples were collected by a gray-scale camera at 200 fps in controlled lighting conditions to prevent flickering. Unlike previous datasets that lack ethnic diversity, the participants are from 13 different ethnicities.

%During data collection,  participants were required to fill a questionnaire, before experiments. The experiment conductor displayed videos that are relevant to the answer of the questionnaire.  %SAMM is coded using the Facial Action Coding System. It includes the ME emotion classes \textit{happy}, \textit{sad}, \textit{surprise}, \textit{angry}, \textit{disgust}, \textit{fear}, \textit{contempt}, and \textit{other}.

%For example, if a participant mentioned the fear of heights, a video of a bungee jump was displayed to induce fear in the targeted participant. All recorded videos are FACS coded with less emphasis on emotional labelling.

\textbf{MEVIEW} \cite{artnerspotting} is  in-the-wild ME dataset. The samples in MEVIEW are collected from poker games and TV interviews on the Internet.
  In total, 31 videos from 16 individuals were annotated in the dataset and the average length of videos is three seconds.   
  %Annotations regarding facial action units and emotional categories are provided.
  %The onset and offset frames of the ME were labeled in long videos.
%During the poker game, players need to hide true emotions which lead to the players under high stake scenario under which MEs likely appear. However, these videos were not recorded for ME study so some valuable moments might have been cut during the TV show post production and the movement of cameras.

\textbf{MMEW}~\cite{ben2021video} contains 300 ME and 900 macro-expression samples acted out by the same participants with a larger resolution ($1920\times 1080$ pixels). MEs and macro-expressions in MMEW were annotated to the same  emotion classes. %(\textit{Happiness, Anger, Surprise, Disgust, Fear, Sadness}).

\textbf{The composite dataset}~\cite{see2019megc} is proposed by the 2nd Micro-Expression Grand Challenge (MEGC2019). The composite dataset merges samples from three spontaneous facial ME datasets: CASME II \cite{Yan2014CASME}, SAMM \cite{davison2018samm}, and  SMIC-HS \cite{li2013spontaneous}.  This is to facilitate the evaluation of newly developed methods. As the annotations in the three datasets vary hugely, the composite dataset unifies emotion labels in all three datasets. The emotion labels are re-annotated as \textit{positive}, \textit{negative}, and \textit{surprise}. 

%

%Consequently, this consolidation includes 442 samples (145 from CASME II, 133 from SAMM, and 164 from SMIC) from 68 subjects (24 from CASME II, 28 from SAMM, and 16 from SMIC). 

\textbf{The compound micro-expression dataset (CMED)}~\cite{zhao2019convolutional,zhao2020compound} is constructed by combining MEs from the CASME, CASME II, CAS(ME)$^2$, SMIC-HS, and SAMM datasets. Specifically, the MEs are divided into basic and compound emotional categories, as shown in Table \ref{dataset}. Psychological studies demonstrate that there are usually complex expressions in daily life. Multiple emotions co-exist in one FE, termed as ``compound expressions" \cite{zhao2020compound}.   Compound expression analysis reflects more complex mental states and more abundant human facial emotions. 

%s: \textit{Negative}, \textit{Positive}, and \textit{Surprised}, and compound emotional categories:  \textit{Positively Surprised}, \textit{Negatively Surprised}, \textit{Positively Negative}, and \textit{Negatively Negative}
 %For instance, angrily  surprise and happily surprise are two distinct compound expressions. 
% \textit{Negative} (233), \textit{Positive} (82), and \textit{Surprised} (70), and compound emotional categories:  \textit{Positively Surprised} (74), \textit{Negatively Surprised} (236), \textit{Positively Negative} (197), and \textit{Negatively Negative} (158).

%\textbf{Synthetic Data}
%Considering the difficulties of fine-labeled ME dataset, we propose a novel synthetic ME dataset based on current manual annotated ME datasets, i.e., CASME II and SAMM, contains over 10,000 sequences with balanced classes. We sincerely welcome the latest efforts and research advances from the scientific community to join this challenge in tackling the micro-expression recognition challenge on our published dataset!

The specific comparisons of the ME datasets are shown in Table \ref{dataset}  and example samples are
shown in Figure \ref{fig:microdatasets}. Although MEVIEW collects MEs in the wild, the number of ME samples is too small to learn robust ME features.  The state-of-the-art approaches are commonly tested on the SMIC-HS, CASME \cite{Yan2013CASME}, CASME II \cite{Yan2014CASME}, and SAMM databases. As some emotions are difficult to trigger, such as fear and contempt, these categories have only a few samples and are not enough for learning. In most practical experiments, only the emotion categories with more than 10 samples are considered. Recently, the composite dataset  is popular, because it can verify the generalization ability of the method on datasets with different natures. For further increasing the MER performance, MMEW collected micro- and macro-expressions from the same subjects which
may be helpful for further cross-modal research.
%inspire promising future research to explore the correlation between micro- and macro-expressions. 

%Following the previous works \cite{Li2013A,Huang2016Spontaneous,Li2015Towards}, only the emotions with more than 10 samples are considered. 
%The emotion categories in CASME are then classified as: \textit{Disgust} (\textbf{D}), \textit{Surprise} (\textbf{S}), \textit{Repression} (\textbf{R}), and \textit{Tense} (\textbf{T}). 
%The samples in CASME II are classified as: \textit{Happiness} (\textbf{H}), \textit{Disgust} (\textbf{D}), \textit{Surprise} (\textbf{S}), \textit{Repression} (\textbf{R}), and \textit{Others} (\textbf{O}). 
%The samples in SAMM are classified as: \textit{Happiness} (\textbf{H}), \textit{Anger} (\textbf{A}), \textit{Surprise} (\textbf{S}), \textit{Contempt} (\textbf{C}), and \textit{Other} (\textbf{O}). 
%All ME samples in the SMIC database are used for experimentation, classified as: \textit{Negative} (\textbf{N}), \textit{Positive} (\textbf{P}), and \textit{Surprise} (\textbf{S}).

%\begin{figure}
%  \centering
%\includegraphics[width=\columnwidth]{image/MEframework.jpg}
%\caption{\textcolor{black}{Deep MER framework.}}
%\label{fig:framework}
%\end{figure}

%\section{Facial action coding system}

\begin{figure*}
     \centering
     \begin{subfigure}[b]{0.78\textwidth}
         \begin{tikzpicture}[scale=0.64]
\tikzset{grow'=right,level distance=88pt}
\tikzset{execute at begin node=\strut}
\tikzset{every tree node/.style={anchor=base west}}

\tikzset{edge from parent/.style=
{draw,
edge from parent path={(\tikzparentnode.east)
--+(+8pt,+0pt)|-
(\tikzchildnode)}}}
\Tree [.{\makecell[c]{MER\\with DL}} [.Input  [.Pre-processing [.Face~detection:~\textit{Adaboost~\cite{viola2001robust}~CNN-based~\cite{matsugu2003subject}} ] [.Face~registration:~\textit{ASM~\cite{cootes1995active}~AMM\cite{cootes2001active}~CNN-based~\cite{zhang2016joint}}  ] [.Temporal~normalization:~\textit{TIM~\cite{zhou2011towards}~CNN-based\cite{ niklaus2018context}}  ]  [.Motion~magnification:~\textit{EVM~\cite{wu2012eulerian}~GLMM\cite{cootes2001active}~{CNN-based}~\cite{oh2018learning}} ] [.Regions~of~interest:~\textit{Grid~\cite{Zhao2007LBP}~FACS\cite{davison2018objective,merghani2020adaptive}~EyeMouth~\cite{shreve2014automatic}~landmarks~\cite{xie2020assisted}~Learning-based~\cite{li2020joint} } ]  [.Data~augmentation:~\textit{Multi-ratio~\cite{xia2019spatiotemporal}~Temporal\cite{li2021micro}~{GAN}~\cite{xie2020assisted,yu2020ice}} ]]
[.Modality [.Static [.Apex~spotting:{\makecell[l]{~\textit{Optical~flow~\cite{patel2015spatiotemporal}~Feature~contrast~\cite{Li2015Towards,han2018cfd}~Frequency~\cite{li2018can}}\\~\textit{CNN-based~\cite{zhang2018smeconvnet,tran2021micro,pan2020local}}}} ]
[.Apex~based~recognition:~\textit{\cite{li2020joint,li2018can,van2019capsulenet,sun2020dynamic}} ] ] [.Dynamic [.Sequence:~\textit{\cite{li2021micro, zhi2019combining, xie2020assisted, kim2016micro,li2018micro}} ] [.Frame~aggregation:~\textit{Onset-Apex~\cite{liong2017micro}~Snippets~\cite{liu2020sma}~Selected~frames~\cite{Kumar_2021_CVPR}}  ] [.Image~with~dynamic~information:~\textit{Dynamic~image~\cite{nie2021geme,verma2019learnet}~Active~image~\cite{verma2020non}} ] [.Optical~flow:{\makecell[l]{~\textit{~Lucas-Kanade~\cite{lucas1986generalized}~Farnebäck~\cite{farneback2003two}~TV-L1\cite{xia2019spatiotemporal}\cite{zhou2019dual}}~\textit{FlowNet~\cite{dosovitskiy2015flownet}}}} ] ] [.Combination:{\makecell[l]{~\textit{{Optical~flow+Apex}~\cite{liu2020offset}~{Optical~flow+Sequence}\cite{sun2019two}~ {Optical~flow+Key~frames}\cite{kim2016micro}   }\\~\textit{{Optical~flow+Landmarks}\cite{kumar2021micro}}}} ] ] ]
[.Network
 [.Block:{\makecell[l]{~\textit{~RES\cite{liu2020multi,nie2021geme}~Inception~\cite{zhou2019dual}~HyFeat \cite{verma2020non}~Capsule\cite{liu2020offset}~RCN \cite{xia2019spatiotemporal}~{Attention}\cite{chen2020spatiotemporal,gajjala2020meranet,li2021micro}}~\textit{{Graph}\cite{lei2020novel,lo2020mer,zhou2020objective}}\\~\textit{{Transformer}\cite{Lei_2021_CVPR}} }} ]
[.Architecture [.Single~stream:~\textit{~2D~\cite{li2018can,le2020dynamic,liu2020sma}~3D~\cite{wang2020eulerian,gajjala2020meranet,chen2020spatiotemporal}} ]
[.Multiple~stream: [.Same~block:~\textit{~Dual~\cite{gan2019off,khor2019dual,yan2020micro}~Triple~\cite{li2019three,song2019recognizing,li2019micro,yang2019merta}~Four~\cite{she2020micro}} ]  [.Different~blocks:~\textit{~Dual~\cite{wang2020micro,peng2019novel,huang2020shcfnet}~Triplet~\cite{liong2019shallow,wu2021tsnn} } ]  [.Handcraft+CNN:~\textit{~Dual~\cite{takalkar2020manifold, hu2018multi,pan2020hierarchical}} ]   ]
[.Cascade:~\textit{CNN+RNN~\cite{nistor2020multi}~CNN+LSTM~\cite{khor2018enriched,zhang2020micro,bai2020investigating,choi2020facial,huang2020shcfnet,zhi2019facial}~\textcolor{black}{CNN+GCN}\cite{xie2020assisted} } ]
[.Multi-task~learning:~\textit{landmark\cite{li2019facial}~Gender\cite{nie2021geme}~{AU}\cite{zhou2020objective}~Multi-binary-class\cite{hu2018multi}} ]
]
[.Training~strategy:~{~\textit{Finetune\cite{chen2020spatiotemporal,liu2020sma}~Knowledge~distillation\cite{hinton2015distilling,komodakis2017paying,sun2020dynamic}}~\textit{Domain~adaption\cite{liu2019neural,zhou2019cross,xia2020learning}}} ]
[.Loss:~\textit{Cross-entropy\cite{kline2005revisiting}~Metric\cite{hadsell2006dimensionality,schroff2015facenet}~Margin\cite{liu2016large,deng2019arcface,wang2018cosface}~{Imbalance} \cite{lin2017focal,zhou2020objective}}  ]
] ]

\end{tikzpicture}
         \label{fig:demonstration}
     \end{subfigure}
     \hfill
     \begin{subfigure}[b]{0.16\textwidth}
 \includegraphics[width=\textwidth]{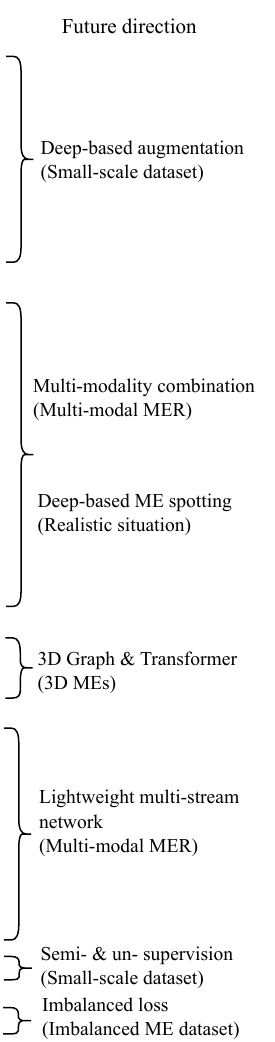}
         \label{fig:real}
     \end{subfigure}
     \hfill
        \caption{Taxonomy for MER based on deep learning. \textcolor{black}{The studies cited on the branches are example approaches discussed in this paper. The future directions and corresponding approaches are shown on the right side. The future directions are annotated in brackets.}}
        \label{fig:Taxonomy}        
\end{figure*}

\section{A taxonomy for MER based on DL}
\label{sec:taxonomy}
Fig. 3 shows a taxonomy  we summarize for MER  based on DL, built along the important components including input and network.   As the ME sequences have subtle movements and limited samples, different inputs have big impacts on MER performance. Thus the input plays an important role in MER. Firstly, the inputs need to be pre-processed for training a robust network. The specific pre-processing approaches and the strengths and shortcomings of various input modalities are discussed in Section  \ref{sec:inputsall}. Then, the networks introduced in Section \ref{sec:network} are utilized to discriminate between MEs. 
\textcolor{black}{A common MER network can be described from four aspects: block, architecture, training strategy, and loss. Firstly, we introduce the special blocks designed to solve the ME challenges. Then, we describe the architecture in terms of single-stream, multi-stream, cascaded networks, and multi-task learning. Finally, the training strategies and loss functions for training networks are discussed. }\textcolor{black}{The future directions are annotated on the right side of Fig. 3. All the methods discussed in this survey are face-based MER with DL.}

%The general MER pipeline is shown in Fig.~\ref{fig:framework}.
% \textcolor{black}{In practical, the neural networks are trained on the training dataset in advance. During testing, the ME samples are directly input to the trained neural networks and get predictions.} 
%In following sections, we discuss the widely used algorithms for each step and recommend the excellent implementations in the light of referenced papers.

\section{Inputs}
\label{sec:inputsall}

\subsection{Pre-processing}
\label{subsec:processing}

Like ordinary FEs, pre-processing involving face detection and alignment is required for robust MER. Compared with common FEs, MEs have low intensity, short duration, and small-scale datasets making MER more difficult. Therefore, besides 
traditional pre-processing steps, motion magnification, temporal normalization, regions-of-interest, and data augmentation have also been undertaken for better MER performance. 

%The developed MER systems need to consider multiple factors and parameters.

%Therefore, before training the deep neural network to learn meaningful features, pre-processing is usually required to align and normalize the visual semantic information conveyed by the face.

%After normalization of detected faces in face registration, noises are filtered and available features are strengthened for better performance. Methods for processing usually involve magnification of subtle features [21] and temporal normalization [33], while noise removal is done using a Wiener [32] or Gaussian filter [34].

\subsubsection{Face detection and registration}
For processing MEs,  face detection which removes the background and gets the facial region is the first step.  One of the most widely used algorithms for face detection is Viola-Jones \cite{viola2001robust} based on a cascade of weak classifiers. However, this method can not deal with large pose variations and occlusions.  Matsugu~\etal~ \cite{matsugu2003subject} firstly adopted CNN network for face detection with a rule-based algorithm,  which is robust to translation, scale, and pose.  Recently, face detectors based on DL have been utilized in popular open source libraries, such as dlib and OpenCV. 
  
%Furthermore, various face detection methods based on DL have been presented to overcome pose variations \cite{sun2013deep, ranjan2017hyperface}.  

%Some methods \cite{wu2004fast} solve these problems by utilizing pose-specific detectors and  probabilistic approaches.

Since spontaneous MEs involve muscle movements of low intensity, even little pose variations and movements may heavily affect MER performance. 
To this end, face registration is crucial for MER. It aligns the detected faces onto a reference face to handle varying head-pose issues for successful MER.
Currently,  one of the most used facial registration methods is the Active Shape Models (ASM) \cite{cootes1995active}  encoding both geometry and intensity information.  Then, the Active Appearance Models (AAM) \cite{cootes2001active} is presented for matching any face with any expression rapidly. 
With the fast development of DL, deep networks with cascaded regression \cite{zhang2016joint} have become the state-of-the-art methods for face alignment due to their excellent performances. %Specifically, in practice, for ME clips,  all frames in the same video clip are registered with the same transformation, as the head movements in micro-expression clips is small. 
%\textcolor{black}{Table 2 investigates facial landmark detection algorithms widely used in deep FER and compares them in terms of efficiency and performance.}
 %registration
 
 %Discriminative Response Map Fitting (DRMF) is a regression approach which can handle occlusions in dynamic backgrounds with lower computational consumption in real time \cite{asthana2013robust}.

\subsubsection{Motion magnification}
%Since the low intensity of micro-expressions leads to difficult micro-expression recognition, motion magnification is important for  recognizable FEs. 

One challenge for MER is that the facial movements of MEs are too subtle to be distinguished. Therefore, motion magnification is important to enhance the ME intensity level. One of the commonly used  methods is  the Eulerian Video Magnification method (EVM)  \cite{wu2012eulerian}. For MEs, the EVM is applied for facial motion magnification \cite{Li2015Towards}. EVM magnifies either motion or color content across two consecutive frames in videos. However, a larger motion amplification level leads to a larger scale of motion amplification, which causes bigger artifacts. Different from EVM considering local magnification, Global Lagrangian Motion Magnification (GLMM) \cite{le2018micro} was proposed for consistently tracking and exaggerating the FEs and global displacements across a whole video. Furthermore, the learning-based motion magnification \cite{oh2018learning} was firstly used in ME magnification by Lei~\etal~ \cite{lei2020novel} through extracting shape representations from the intermediate layers of networks. Compared with the traditional methods, the shape representations from the intermediate layers introduce less noise.

%\cite{oh2018learning，  le2019seeing，chen2020deepmag， takeda2019video，shabi2020motion，lei2019facial}

%As the proposal takes an opposite approach to a previous pivotal work, i.e. local Amplitude-based Eulerian Motion Magnification (AEMM). GLMM and AEMM are theoretically analyzed for potential advantages and disadvantages, especially with respect to how magnified noise and distortions are dealt with. Then, both GLMM and AEMM are empirically evaluated and compared using the CASME II micro-expression corpus

%α is a parameter that controls the level of motion amplification. Bigger values of α lead to larger scale of motion amplification, but also can cause bigger displacement and artifacts. 

\subsubsection{Temporal Normalization (TN)}

Besides the low intensity,  the short and varied duration also increases the difficulty for robust MER. This problem is especially serious when the videos are filmed with relatively low frame rate. To solve this issue, the Temporal Interpolation Model\cite{zhou2011towards} (TIM) was introduced to interpolate all ME sequences into the same specified length based on path graph between the frames. There are three strengths of applying TIM: 1) up-sampling ME clips with too few frames; 2)  more stable features can be expected with a unified clip length; 3) extending ME clips to long sequences and sub-sampling to short clips for data augmentation. Additionally, CNN-based temporal interpolation  \cite{niklaus2018context} have been proposed to solve complex scenarios in reality.

%For example, some MEs may only last for four frames, when the video is recorded at a standard speed of 25 fps. 
% temporal relations between the frames.
%The TIM method relies on a path graph to characterize the structure of a sequence of frames.

%A sequence-specific mapping is learned to connect frames in the sequence and a curve embedded in the path graph so that the sequence can be projected onto the latter. The curve, which is a continuous and deterministic function of a single variable t in the range of [0,1], governs the temporal relations between the frames. Unseen frames occurring in the continuous process of an ME are also characterized by the curve. Therefore a sequence of frames after interpolation can be generated by controlling the variable t at different time points accordingly.

\begin{figure}
  \centering
%  \centerline{\includegraphics[width=8.5cm]{flowImg.pdf}}
\includegraphics[width=\columnwidth]{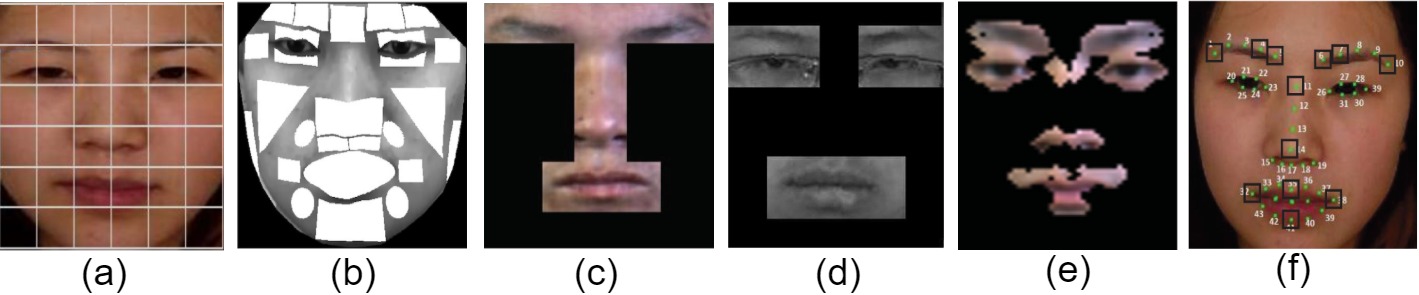}
\caption{Examples of regions of interest. (a) Equal block; (b) FACS-based RoIs \cite{davison2018objective}; (c)  RoIs Masked eye and cheek \cite{le2020dynamic}; (d)  Eye and mouth \cite{liong2018hybrid}; (e) Difference-based ME datasets~\cite{hong2019Spatiotemporal}; (f) landmark-based local regions~\cite{li2018ltp}.}
\label{fig:roi}
\end{figure}

\subsubsection{Regions of interest (RoIs)}
FEs are formulated by basic facial movements \cite{rosenberg2020face,friesen1978facial}, which correspond to specific facial muscles and relate to different facial regions. In other words, not all facial regions contribute equally to FER. Especially for MEs, the MEs only trigger specific small regions, as MEs involve subtle facial movements.  Moreover, the empirical experience and quantitative analysis in~\cite{huang2012towards} found that the outliers such as eyeglass have a seriously negative impact on the performance of MER. Therefore, it is important to suppress the influence of outliers.

Some studies alleviate the influence of regions without useful information by extracting features on the RoIs~\cite{le2020dynamic}. Several MER approaches \cite{Li2015Towards,chen2022block} divided the entire face into several equal blocks for better describing local changes (see Fig.~\ref{fig:roi} (a)).  Davison~\etal~ \cite{davison2018objective,merghani2020adaptive}  selected RoIs from the
face based on the FACS~\cite{friesen1978facial}, shown in  Fig.~\ref{fig:roi} (b). In addition, to eliminate the noise caused by the eye blinking and motion-less regions, Le~\etal \cite{le2020dynamic}  proposed to mask the eye and cheek regions for each image (see Fig.~\ref{fig:roi} (c)). 
However, the motion of eyes has a big contribution to MER under certain situations, \eg~lid tighten refers to negative emotion. In work \cite{liong2018hybrid},  Liong~\etal~ utilized the eyes and mouth regions for MER, as shown in Fig.~\ref{fig:roi} (d). Besides, Xia~\etal~\cite{hong2019Spatiotemporal} found that the regions around the eyes, nose, and mouth are mostly active for MEs and can be chosen as RoIs through analyzing difference heat maps of ME datasets, as shown in Fig.~\ref{fig:roi} (e). Furthermore, Xie~\etal~\cite{xie2020assisted} and Li~\etal~\cite{li2018ltp} proposed to extract features on small facial blocks located by facial landmarks  (see Fig.~\ref{fig:roi} (f)). In this way, the dimension of learning space can be drastically reduced and helpful for deep model learning on small ME datasets. %Above ROIs are shown in Figure. \ref{fig:roi}.

%All above researches are based on the assumption that all the selected  RoIs have an equal contribution to MER. However, in practice, different RoIs come from different MEs.Li~\etal~ \cite{li2020joint} designed a local information learning module to automatically learn the regions contributing most ME information following the concept of multi-instance learning (MIL)~\cite{huang2018fast}. The local information learning module aims to suppress the affect of outliers and improve the discriminative representation ability. 

%For example, anger is mostly related to AU4 (Brow lower) or AU7 (Lids tight)~\cite{TMM2019AU}. Compared with the motionless regions, the local region related to AUs may contribute more information to MER. 

%Owing to the  subtle, fast and involuntary characteristics of MEs, it is difficult to collect MEs.
\subsubsection{Data augmentation}
The main challenge for MER with DL is the  small-scale ME datasets.  The current ME datasets are too limited to train a robust DL model from scratch, therefore data augmentation is necessary. The common way for data augmentation is random crop and rotation in terms of the spatial domain. Xia~\etal~augmented MEs through magnifying MEs with multiple ratios \cite{xia2019spatiotemporal}. Fig.~\ref{fig:Data_augmentation} (a) and (b) show the examples of magnified ME apex frames with different ratios on the basis of EVM \cite{wu2012eulerian} and learning-based magnification \cite{oh2018learning}, respectively. 
 Additionally, Generative Adversarial Network (GAN) \cite{zhang2018joint} can augment data by producing synthetic images.  Xie~\etal~\cite{xie2020assisted} introduced the AU Intensity Controllable GAN (AU-ICGAN) to synthesize subtle MEs. As Fig.~\ref{fig:Data_augmentation} (d) shows, the ME sequences with continuous AU intensity can be synthesized through~\cite{xie2020assisted}. Yu~\etal~\cite{yu2020ice} proposed a Identity-aware and Capsule-Enhanced Generative Adversarial Network (ICE-GAN) to complete the ME synthesis and recognition tasks. ICE-GAN outperformed the winner of MEGC2019 by 7$\%$, demonstrating the effectiveness of GAN for ME augmentation and recognition.  The synthesized images corresponding to different emotions are shown in Fig.~\ref{fig:Data_augmentation} (c).
Besides, Liong~\etal~\cite{liong2020evaluation} utilized conditional GAN to  generate optical-flow images to improve the MER accuracy based on computed optical flow. For ME clips, sub-sampling MEs from extended ME sequences through TIM can augment ME sequences~\cite{li2021micro}.

%Zhang~\etal~ \cite{zhang2018joint} exploited different poses and expressions jointly through GAN for simultaneous facial image synthesis and pose-invariant FER. Furthermore,Pumarola~\etal~ \cite{pumarola2018ganimation} proposed a novel GAN conditioning scheme based on Action Unit annotations which allows controlling the AU intensity and constructing expressions with various intensities. 

\begin{figure}
  \centering
%  \centerline{\includegraphics[width=8.5cm]{flowImg.pdf}}
\includegraphics[width=\columnwidth]{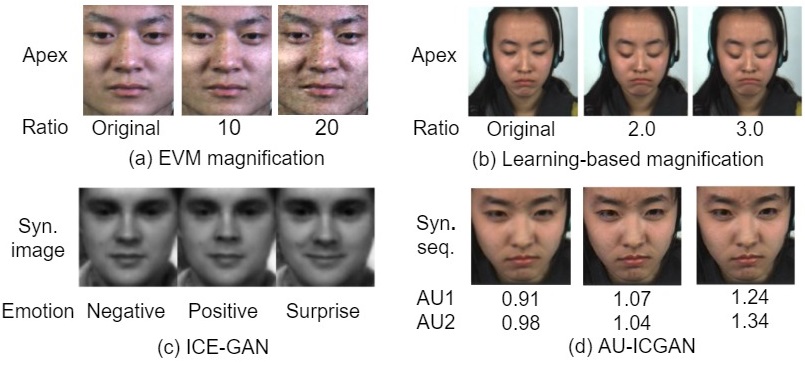}
\caption{Examples of magnified and synthesized MEs.}
\label{fig:Data_augmentation}
\end{figure}

\begin{figure*}
  \centering
%  \centerline{\includegraphics[width=8.5cm]{flowImg.pdf}}
\includegraphics[width=18.2cm]{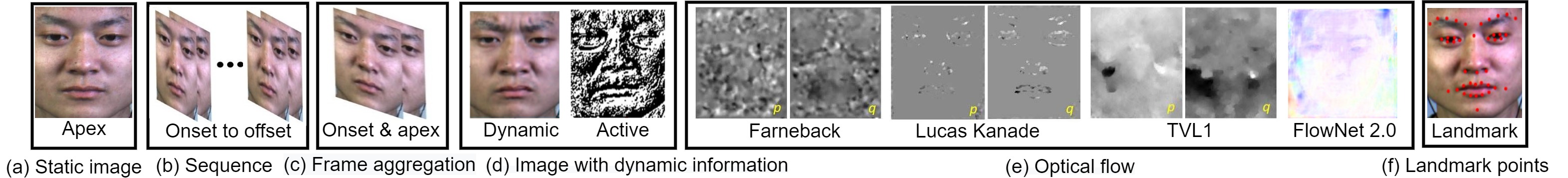}
\caption{Examples of various inputs.}
\label{fig:opticalflow}
\end{figure*}
\subsection{Input modality}
\label{sec:inputs}
Since the MEs have low intensity, short duration, and limited data, it is challenging to recognize MEs based on DL and  the MER performance varies with different inputs.  In this section, we describe the various ME inputs and summarize their strengths and shortcomings, as shown in Table \ref{tab:input}. 

\subsubsection{Static image} 
For FER, a large volume of existing studies are conducted on static images without temporal information due to the availability of the massive facial images online and the convenience of data processing. Inspired by efficient FER with static images, some researchers~\cite{li2018can,van2019capsulenet} explored the MER based on the apex frame with the largest intensity of facial movement among all frames (See Fig.~\ref{fig:opticalflow} (a)). Li~\etal~\cite{li2020joint}  studied the contribution of the apex frame and verified that DL can achieve good MER performance with the single apex frame. Furthermore,  the research of Sun~\etal~ \cite{sun2020dynamic} showed that the apex frame-based methods can effectively utilize the massive static images in macro-expression databases \cite{sun2020dynamic} and obtain better performance than onset-apex-offset sequences and the whole videos.

%Multiple apex frame spotting approaches were proposed to locate the apex frame in ME sequences for MER based on apex frames.
Apex spotting is one of the key components for building a robust MER system based on apex frames. Patel~\etal~\cite{patel2015spatiotemporal} computed the motion amplitude of optical flow shifted over time to locate the onset, apex, and offset frames of MEs, while other works \cite{Li2015Towards,han2018cfd} exploited feature differences to detect MEs in long videos.  However, optical flow-based approaches required complicated feature operation and the feature contrast-based methods ignored ME dynamic information. Different from above methods estimating the facial change in the spatio-temporal domain, Li~\etal~\cite{li2018can} proposed to  locate the apex frame in rapid ME clips through exploring the information in the frequency domain which clearly describes  the  rate  of  change. Furthermore, SMEConvNet \cite{zhang2018smeconvnet} firstly adopted CNN for ME spotting and a feature matrix processing was proposed for locating the apex frame in long videos. Following SMEConvNet, various CNN-based ME spotting methods \cite{tran2021micro,pan2020local} were proposed. In general, the performance of CNN-based spotting method is limited because of the small-scale ME datasets and mixed macro- and micro-expressions clips in long videos. Further studies on reliable spotting methods are required in the future. 

%All above ME spotting methods estimated the facial change in the spatio-temporal domain. However, the temporal changes of MEs are not salient, due to the low intensity and short duration characteristics. Thus, it is tough to detect the apex frame in spatio-temporal domain. 

\subsubsection{Dynamic image sequence} 

As the facial movements are subtle in the spatial domain, while change fast in the temporal domain, the temporal dynamics along the video sequences are essential in improving the MER performance. In this subsection, we  describe the various dynamic inputs.

\textit{Sequence.} Most ME researches  utilize consecutive frames in video clips \cite{TIPcolor2018,  Zhao2007LBP,mayya2016combining}, as shown in~Fig. \ref{fig:opticalflow} (b). 
 With the success of 3D CNN \cite{ji20123d} and Recurrent Neural Network (RNN)~\cite{medsker2001recurrent} in video analysis \cite{ullah2017action,yang2019asymmetric},  MER based on sequence \cite{wu2021tsnn, li2021micro, zhi2019combining, xie2020assisted, kim2016micro,li2018micro, peng2020recognizing} is developed that considers the spatial and temporal information simultaneously. However, the computation  cost is relatively high and the complex model tends to overfit the small-scale training data.

\textit{Frame aggregation.} 
MEs are mostly collected with a high-speed camera (\eg~200 fps) to capture the rapid subtle changes. Liong~\etal~discovered that there is redundant information in ME clips recorded with high-speed cameras~\cite{Liong2017Less}. The redundancy could decrease the performance of MER. The experimental results of~\cite{Liong2017Less} demonstrate that the onset, apex, and offset frames can provide enough spatial and temporal information to ME classification.  Liong~\etal~\cite{liong2017micro} extracted features on onset and apex frames for MER, as shown in~ Fig. \ref{fig:opticalflow} (c). Furthermore, in order to avoid apex frame spotting, Liu~\etal~\cite{liu2020sma} and Kumar~\etal~\cite{Kumar_2021_CVPR} designed  simple strategies to select aggregated frames automatically.

%SMA-STN~\cite{liu2020sma} designed a Dynamic Segmented Sparse Imaging  (DSSI)  module to generate  snippets through sparsely sampling three frames from evenly divided sequences and LR-GACNN \cite{Kumar_2021_CVPR} selected frames based on the magnitude of the optical flow. In this way, the subtle movement changes and significant segments in a ME sequence can be efficiently captured  for robust MER. 

\textit{Image with dynamic information.}
Image with dynamic information~\cite{bilen2016dynamic} is a standard image that holds the dynamics of an entire video sequence in a single instance. The dynamic image generated  by using the rank pooling algorithm has been successfully used in MER~\cite{nie2021geme,le2020dynamic,verma2019learnet,bilen2017action} to summarize the subtle dynamics and appearance in an image.  Similar to dynamic images, active images \cite{verma2020non} encapsulated the spatial and temporal information of a video sequence into a single instance through estimating and accumulating the change of each pixel component (See Fig.~\ref{fig:opticalflow} (d)). 

%Furthermore, Liu~\etal~ \cite{liu2020sma} designed a dynamic segmented sparse imaging module (DSSI) to compute a set of dynamic images as the input data of subsequent models.

\begin{table*}
\renewcommand{\arraystretch}{1.3}
\centering
\small
\caption{The comparisons of inputs for MER}

\scalebox{0.84}{ 
\begin{tabular}{|l|l|l|l|l|}
 \bottomrule[1.5pt]
% \bottomrule[1.5pt]    
\multicolumn{2}{|c|}{Input modality}& \multicolumn{1}{c|}{Strength}&\multicolumn{1}{c|}{Shortcoming} \\ \bottomrule[1.5pt]
\multicolumn{2}{|c|}{Static} &Efficient; Take advantage of  massive facial images  & \makecell[l]{Require magnification and apex detection \\  Without temporal information}
\\ \bottomrule[1.5pt]
\multirow{4}{*}{Dynamic}&Sequence&Process directly&Not efficient; Information  redundancy\\  \cline{2-4}
&Frame  aggregation&Efficiently leverage key temporal information & Require apex  detection\\\cline{2-4}
&Image with dynamic information& Efficiently embed  spatio-temporal information    & Require dynamic information computation
\\  \cline{2-4} 
&Optical flow&Remove identity to some degree; Movement considered & Optical flow computation is necessary\\ \bottomrule[1.5pt]
\multicolumn{2}{|c|}{Combination}& Explore spatial and temporal information&High computation cost\\ \bottomrule[1.5pt]
\end{tabular}
}
\\
 \label{tab:input}
\end{table*}

\textit{Optical flow.} The motion between ME frames contributes important information for ME recognition.  Optical flow approximates the local image motion, which has been verified to be helpful for motion representation~\cite{barron1994performance}. It specifies the magnitude and direction of pixel motion in a given sequence of images with a two-dimension vector field (horizontal and vertical optical flows).  In recent years, several novel methodologies
 have been presented to improve optical flow techniques~\cite{lucas1986generalized,horn1981determining,senst2012robust,farneback2003two,wedel2009improved}, such as Farnebäck’s~\cite{farneback2003two}, Lucas-Kanade~\cite{lucas1986generalized},  TV-L1~\cite{wedel2009improved}, FlowNet~\cite{dosovitskiy2015flownet}, as shown in Fig. 
\ref{fig:opticalflow} (e). 
\textcolor{black}{Currently, many MER approaches utilize optical flow to represent the micro-facial movement and reduce the identity characteristic~\cite{xia2019spatiotemporal,song2019recognizing,allaert2019optical}.  Researches~\cite{xia2019spatiotemporal,song2019recognizing} indicated  that  optical  flow-based  methods  always  outperform appearance-based methods. To further capture the subtle facial changes, multiple works~\cite{zhou2019dual,liong2019shallow,zhou2020objective} extracted features on computed optical flows on the onset and mid-frame/apex in horizontal and vertical directions separately. 
}

%Different from above studies, Belaiche~\etal~ proposed a compact optical flow representation embedding motion direction and magnitude into one matrix \cite{belaiche2020cost}. Their results suggested that the optical flow direction, especially vertical direction, contributes important information for ME recognition. Besides, Su~\etal~\cite{su2021key} analyzed both the first- and second-order motion of optical flow to further improve the MER performance.

%(STSTNET)~\cite{liong2019shallow}  and concatenated optical strain, horizontal and vertical optical flow as input data using onset and apex frame.
 
\subsubsection{Input combination}
Considering the strengths of apex frame and dynamic image sequences, some works \cite{song2019recognizing,kim2016micro,liu2020offset,sun2019two} analyze multiple inputs to  learn features from different cues in ME videos.
Specifically, in Liu~\etal's work \cite{liu2020offset}, the apex frames and optical flow are utilized to extract static-spatial and temporal features, respectively. Besides the above modalities, Song~\etal~\cite{song2019recognizing} added local facial regions of the apex frame as inputs to embed the relationship of individual facial regions for increasing the robustness of MER.   In addition, Sun~\etal~\cite{sun2019two} employed optical flow  and sequences for fully exploring the temporal ME information. 
Recently, inspired by the successful application of landmarks in facial analysis (See Fig. \ref{fig:opticalflow} (f)), Kumar~\etal~\cite{kumar2021micro} proposed to fuse the landmark graph and optical flow to enhance the discriminative ability of ME repression. Currently, the approaches with multiple inputs achieve the best MER performance through leveraging as much as ME information on limited ME datasets.

%Moreover, Kim~\etal~\cite{kim2016micro} encoded sequences with spatial features at key expression-states to increase the expression class separability of the learned ME feature.

%Song~\etal~ \cite{song2019recognizing} proposed to learn the static-spatial and temporal features from apex frame, local regions of apex frame and optical flow between onset and apex frames to improve MER performance, respectively.

%The Optical flow fields between three frames (the onset, apex,and offset frames). a Some of recent studies [8], [27], [28] showed that the apex frame contains more ME-aware information, and thus we design a static-spatial stream CNN in the TSCNN to learn the static-spatial feature from the gray image of the apex frame CNN is mainly inspired by recent findings in [9], [16], [29], [30]. Their studies have proved that the facial local region information has indeed contributions to distinguishing different MEs. Finally, following some studies in [19], [24], [25], [31]

\subsection{Discussion}
In summary, the input is one of the key components to guarantee robust MER. 
The various ME inputs have different strengths and shortcomings. The comparisons of inputs are shown in Table \ref{tab:input}.

The input pre-processing is the first step in the MER system. Besides common face pre-processing approaches (face detection and registration), motion magnification, RoIs, and TIM also play important roles for robust MER, due to the subtle and rapid characteristics of MEs. Current motion magnification approaches always introduce noises and artifacts. More effective  motion magnification approaches should be explored. Furthermore, considering the small-scale ME datasets are far from enough to train a robust deep model, data augmentation is necessary for MER. In the future, studying more robust GAN-based ME generation approaches is a promising research direction.

Regarding the static input, the apex-based MER can reduce computational complexity and take advantage of the massive FEs to resolve the small-dataset issue in some degree. But, magnification is necessary since all the temporal information is dropped in single apex-based methods and the motion intensity is still low in the apex frames. Moreover, as the apex label is absent in some ME datasets, the performance of apex-based MER severely relies on the apex detection algorithm. Currently, the apex frame detection in long videos is still challenging. The end-to-end framework for apex frame detection and MER needs to be further studied.

Compared with the static image, the dynamic input is able to leverage spatial and temporal information for robust MER. The \textcolor{black}{simplest} dynamic input is ME sequence which doesn't require extra operations.  However, there is redundancy in ME sequences, and the complexity of the deep model is relatively high and tends to overfit on small-scale ME datasets.  To solve the problem of redundancy, frame aggregation cascading multiple key frames is  utilized. 
Besides, the dynamic image improves the computation efficiency through embedding  the temporal and spatial information to a still image. It can simultaneously consider spatial and temporal information in one image without challenging apex frame detection. Furthermore, optical flow is widely used for MER as the optical flow describes the motions and removes the identity in some degree. However, most of the current optical flow-based MER methods are based on traditional optical flow, which is not end-to-end. In the future, more DL-based optical flow extraction can be further researched.

In addition, combining various inputs is the inevitable trend to fully explore spatial and temporal information and leverage the merits of various inputs. Correspondingly, the combined inputs also inherit the shortcomings of the inputs. However, the multiple inputs could be
 complementary in some degree. So far, the method with various inputs has achieved the best performance. Considering the success of multiple inputs and limited ME samples, more combined modalities, such as optical flow, key frames, and landmarks can be promising research directions.

\begin{figure*}
  \centering
%  \centerline{\includegraphics[width=8.5cm]{flowImg.pdf}}
\includegraphics[width=18.5cm]{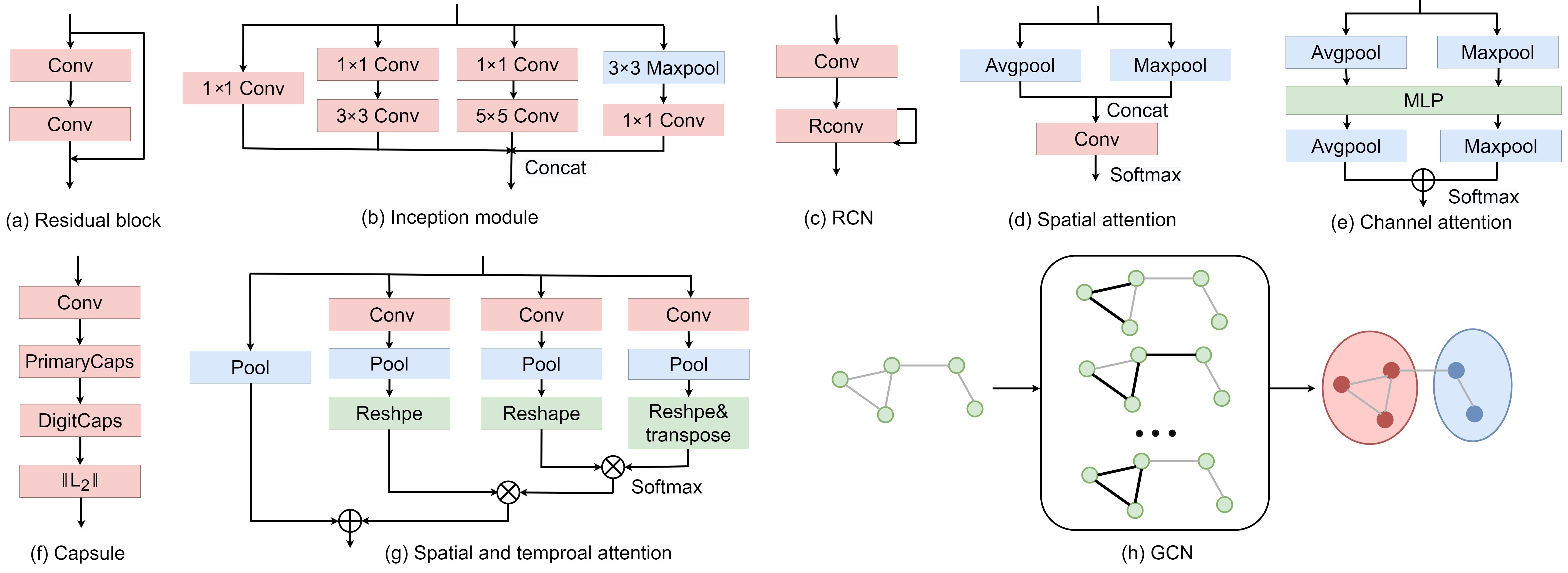}
\caption{Special blocks: (a) Residual block \cite{he2016deep}  (b) Inception module\cite{szegedy2017inception}; (c) RCN \cite{xia2019spatiotemporal};    (d) Spatial attention of CBAM \cite{chen2020spatiotemporal} (e) Channel attention of CBAM \cite{chen2020spatiotemporal};  (f) Capsule module \cite{sabour2017dynamic};  (g) Spatio-temporal attention \cite{wang2020eulerian};  (h) GCN \cite{verma2020non}.}
\label{fig:special}
\end{figure*}

\section{Deep networks for MER} 
\label{sec:network}
Convolutional Neural Networks (CNNs) have shown excellent performances for various computer vision tasks, such as action recognition \cite{pareek2021survey} and FER \cite{li2020deep}. In general, for image classification,  CNNs employ two dimensional convolutional kernels (denoted as 2D CNN) to leverage spatial context across the height and width of the images to make predictions. Compared with 2D CNN, 
CNNs with three-dimensional convolutional kernels (denoted as 3D CNN) are verified more effective for exploring spatio-temporal information of videos \cite{hara2018can}. 3D CNN can take advantage of spatio-temporal information  to improve the performance but comes with a computational cost because of the increased number of parameters. Moreover, the 3D CNN only can deal with videos with the fixed length due to the pre-defined kernels. Recurrent Neural Network (RNN) \cite{medsker2001recurrent} 
was proposed to process the time series data  with various duration. Furthermore,  Long Short-Term Memory (LSTM) was developed to settle the vanishing gradient problem that can be encountered when training RNNs. 

Unlike common video-based classification problems, for the recognition of subtle, fleeting, and involuntary MEs, 
various DL approaches have been proposed to boost MER performance. In this section, we introduced  the approaches in the view of special blocks, network architecture, training strategy, and loss.

%In this section, we gives a review of DL-based MER. Firstly, we introduce the three major aspects of neural networks, including the blocks, structure, and loss. A review of graph network based MER is followed. Finally, we discuss the multi-task and transfer learning utilization in MER.

\subsection{Network block}

In terms of solving the two main ME challenges: overfitting on small-scale ME datasets and low intensity of MEs, various effective network blocks have been utilized and designed, such as ResNet family with residual modules \cite{he2016deep,xie2017aggregated,wu2019wider}, and Inception module \cite{szegedy2017inception}. In this subsection, we introduce the special network blocks utilized for MER improvement.
%With the developments of DL, various efficient network blocks have been proposed, such as ResNet family with residual modules \cite{he2016deep,xie2017aggregated,wu2019wider}, and Inception module \cite{szegedy2017inception}. In this subsection, we introduce the special network blocks utilized for MER improvement. 

  For the challenge of small-scale datasets, recent researches~\cite{he2016deep} demonstrate that  residual blocks with shortcut connections (shown in Fig.~\ref{fig:special} (a)) achieves easy optimization and  reduces the effect of the vanishing gradient problem. Multiple MER works~\cite{peng2019novel,belaiche2020cost,wen2020cross,lei2020novel,lai2020real} employed residual blocks for robust recognition on small-scale ME datasets.  Instead of directly applying the shortcut connection,~\cite{chinnappa2021residual} further designed a convolutionable shortcut to learn the important residual information and AffectiveNet~\cite{verma2020affectivenet} introduced an MFL module learning the low- and high-level feature parallelly to increase the discriminative capability between the inter and intra-class variations.

Since the fully connected layer requires lots of parameters which makes it prone to extreme loss explosion and overfitting~\cite{li2017action},  the Inception module~\cite{szegedy2015going}  aggregates different sizes of filters to compute multi-scale spatial information and assembles $1\times1\times1$ convolutional filters to reduce the dimension and parameter, as shown in Fig.~\ref{fig:special} (b). Multiple works~\cite{zhou2019dual,zhou2020objective} utilized  the Inception module for efficient MER. Inspired by the Inception structure, a Hybrid Feature (HyFeat) block \cite{verma2019hinet,verma2019learnet,verma2020non} was proposed to preserve the domain knowledge features for expressive regions of MEs and enrich features of edge variations through using different scaled convolutional filters.  

Furthermore, considering the fact that CNN with more convolutional layers has stronger representation ability, but easy to overfit on small-scale datasets, paper \cite{xia2019spatiotemporal} and~\cite{xia2020revealing} introduced Recurrent Convolutional Network (RCN) which achieved a shallow architecture though recurrent connections, as shown in Fig.~\ref{fig:special} (c). 

%Moreover, LGCcon learned discriminative ME representations from the RoI contributing major emotion information\cite{li2020joint} though Multiple-instance-leaning (MIL).
%Except for getting discriminative features on RoIs, the relationship of RoIs is very important.  For example, inner brow raiser and outter brow raiser usually exhibits simultaneously to indicate surprise.

%many studies leverage 

On the other hand, MEs perform as the combination of multiple facial movements. The latent semantic information among subtle facial changes contributes important information for MER performance. Recent researches illustrate that the Graph Convolutional Network (GCN) is effective to model these semantic relationships and can be leveraged for face analysis tasks, as shown in Fig.~\ref{fig:special} (h). Inspired by the successful application of GCN in FER, \cite{lei2020novel,xie2020assisted,lo2020mer,zhou2020objective} developed the GCN for MER to further improve the performance by modeling the relationship between the local facial movements.  Lei~\etal~\cite{lei2020novel,Lei_2021_CVPR} built graphs on the RoIs along facial landmarks contributing information to subtle MEs. The TCN residual blocks \cite{bai2018empirical,lei2020novel} and transformer \cite{parmar2018image,Lei_2021_CVPR} were applied for reasoning the relationships of RoIs. On the other hand, as the FE analysis can be benefited from the knowledge of AUs and FACS, the works \cite{xie2020assisted,lo2020mer,zhou2020objective} built graph on AU-level representations to boost the MER performance by inferring the AU relationship. 

%In graph definition, there are two key elements node and edges demonstrating the graph representations and relations.

%AU-feature graph based approaches \cite{xie2020assisted,zhou2020objective} extracted the feature maps representing corresponding AUs and recognized MEs based on the AU graph features \cite{xie2020assisted}.

%For AU-level graph representations, the GCNs can fall into AU-label graphs and AU-feature graphs. For AU-label graphs,the graph is built on the label distribution of training data

%Besides graph,  Hinton~\etal~proposed a Capsule Neural Network (CapsNet) \cite{sabour2017dynamic} to describe the structure of facial attributes through better model hierarchical relationships by routing procedure,  as shown in Fig.~\ref{fig:special} (f). Several MER works~\cite{van2019capsulenet,liu2020offset,yu2020ice} employed CapsNet to explore part-whole relationships on face to promote MER performance.  

Besides graph,   Capsule Neural Network (CapsNet) \cite{sabour2017dynamic} was employed to  explore part-whole relationships on face to promote MER performance through better model hierarchical relationships by routing procedure~\cite{van2019capsulenet,liu2020offset,yu2020ice},  as shown in Fig.~\ref{fig:special} (f). 

%RoI - spatial relationship - spatial and temporal -spatial temporal channel

%Considering the relationship of RoIs is also important, ~\eg~inner brow raiser and outter brow raiser usually exhibits simultaneously to
%indicate surprise, the works \cite{li2021micro} and \cite{wen2020cross} explored covariance correlation of local regions and multi-head self-attention to emphasize the spatial relationships, respectively. 

%Considering special ME characteristics, such as the combination of facial muscle movements, can be utilized for MER.

In addition, since MEs have specific muscular activations on the face, MEs are related with local regional changes~\cite{acharya2018covariance}. Therefore, it is crucial to highlight the representation on RoIs~\cite{xuesongCVPR2019,zhang2020micro}. 
Several approaches \cite{wang2020micro,takalkar2021lgattnet,bai2020detection,hashmi2021larnet,su2021key,wei2022novel} have shown the benefit of enhancing spatial encoding with attention module. 

Except for spatial information, the temporal change also plays an important role for MER. As MEs have rapid changes, the frames have unequal contribution to MER. Wang~\etal~\cite{wang2020eulerian} explored a global spatial and temporal attention module (GAM) based on the non-local network \cite{wang2018non} to encode wider spatial and temporal information to capture local high-level semantic information, as shown in Fig. \ref{fig:special} (g). 
%Moreover, Segmented Movement-Attending Spatio-temporal Network (SMA-STN)  \cite{liu2020sma} utilized a global self-attention module and non-local self-attention block to learn weights for the temporal segments to capture the global context information of sequence and the long-range dependencies of facial movements.

%Squeezeand-and-Excitation network (SEnet) \cite{hu2018squeeze} adaptively recalibrates channel-wise feature responses based on the channel relationship which are regarded as channel attention. Inspired by SEnet,

Moreover,  Yao~\etal~\cite{9103767} learned the weights of each feature channel adaptively through adding squeezeand-and-excitation blocks.  Additionally, recent works \cite{chen2020spatiotemporal,gajjala2020meranet,li2021micro,wang2021micro} encoded the spatio-temporal and channel attention simultaneously to further boost the representational power of MEs. Specifically, 
CBAMNet \cite{chen2020spatiotemporal} presented a convolutional block attention module (CBAM) cascading the spatial attention module (see Fig. \ref{fig:special} (d)) and channel attention module (see Fig. \ref{fig:special} (e)).  %Gajjala~\etal~ \cite{gajjala2020meranet} proposed a 3D residual attention network (MERANet) to re-calibrate the  spatial-temporal and channel features through residual modules.   

\begin{figure*}
  \centering
%  \centerline{\includegraphics[width=8.5cm]{flowImg.pdf}}
\includegraphics[width=18cm]{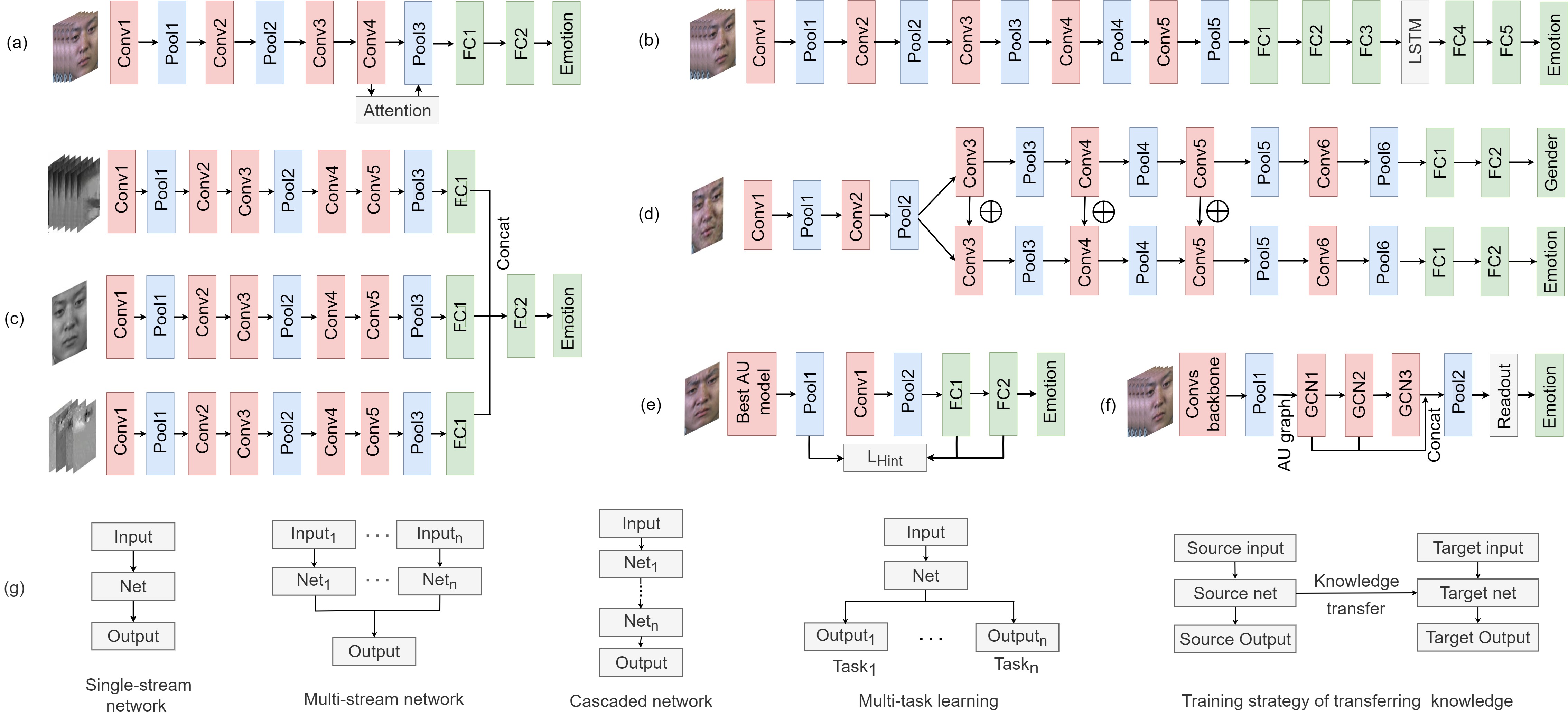}
\caption{(a) GAM based on single stream \cite{wang2020eulerian};  (b)  CNN cascaded with LSTM \cite{bai2020investigating};
(c) TSCNN~\cite{song2019recognizing} based on three-stream network; 
(d) A dual-stream multi-task network incorporating gender detection  designed by GEME \cite{nie2021geme}; (e) CNN cascaded with GCNs on the basis of AU-feature graph   \cite{xie2020assisted}; (f) MER based on knowledge distillation~\cite{sun2020dynamic};  and  (g)  The general concept of single-stream, multi-stream, cascaded networks, multi-task learning, and the training strategy of transferring knowledge. }
\label{fig:classical}
\end{figure*}

In summary, due to the special characteristics of MEs, many DL-based methods designed special blocks to extract discriminative ME representations from the latent semantic information. Recent MER researches indicate that attention and graph blocks are effective to model the semantic relationships. Current GCN-based MER are always based on the local facial regions and AU labels. In the future, more compact and concise representation, such as landmark location, can be further developed for efficient MER. 
Moreover, the transformer \cite{parmar2018image} has been verified effectively on modeling the relationship. For future MER research,  transformers can be further applied to model the relationships between facial landmarks, AUs, RoIs and frames to enhance ME representation.
On the other hand, other special blocks \cite{xia2019spatiotemporal} targeted at learning discriminative ME features with less parameters to avoid overfitting. In the future, more efficient blocks should be studied to dig subtle ME movements on limited ME datasets.

%Wang~\etal~ employ 3DCNN with a global attention module (GAM) to proposed Eulerian motion-based 3D convolution network (EM-C3D).
%LGAttNet \cite{hashmi2021larnet} using dual-stream local and global attentions.

%In order to boost the representational power of a network, 

%Squeezeand-Excitation network (SEnet) focused on the channel relationship and adaptively recalibrates channel-wise feature responses based on the interdependencies between channels.

%\cite{chen2020spatiotemporal}Spatiotemporal Convolutional Neural Network with Convolutional Block Attention Module for MER

%\cite{wang2020micro} Micro-attention for MER

%\cite{gajjala2020meranet} MERANet: Facial MER using 3D Residual Attention Network

%\cite{takalkar2021lgattnet}LGAttNet: Automatic ME detection using dual-stream local and global attentions

\subsection{Network architecture}
%So far, the deep learning methods used in MER include the classical model and graph-based networks. There are a few works about graph-based network, and breakthroughs have been made. the classical models include CNN, RNN and 3DCNN, and this section will introduce it from three aspects: single stream-based, multiple streams-based, and combination-based. CNN, 3DCNN and RNN are the deep learning methods commonly used for image and video analysis. CNN can extract spatial features from image (two dimensions) data, and 3DCNN and CNN+RNN not only can extract spatial features from video data but also extract temporal features from video (three dimensions) data. MEs involve the facial movement over a period of time, and the sample of MEs belong to video data. Thus, 3DCNN and CNN+RNN are suitable to be employed in the field of MER. some researches prove that the apex frame in the ME video has huge contributions, and the onset and apex frame include the main movement of MEs. In addition, some methods can transform the video into one single image that represents the movement information. Thus, the CNN also was widely used to recognize MEs because of low model parameters, simple structures and more effective models. In these works, different blocks, mechanisms, inputs and models are considered to improve performance. The details are as follow:
Besides designing special blocks for discriminative ME representation, the way of combining the blocks is also very important. The current network architecture of MER methods can be classified to five categories: single-stream, multi-stream, cascaded, multi-task learning and transfer learning. In this section, we will discuss the details of the five network architectures.

\subsubsection{Single-stream networks} 
%%limited data - shallow network - Fintune deep - adding attention -3D

%CNN and 3DCNN are commonly used in deep learning, and some works tried to employ single stream-based CNN and 3DCNN models to solve the task of MER. Except for raw frames, apex frame, optical flow and the dynamic images are calculated as input for improving performance. In addition, some mechanisms and blocks were added in basic models, such as attention, shallow models, hybrid and decoupled feature learning.

Typical deep MER methods adopt single CNN with individual input \cite{li2021multi}. The apex frame, optical flow images and dynamic images are common inputs for single-stream 2D CNNs, while single-stream 3D CNNs extract the spatial and temporal features from ME sequences directly. Considering the limited ME samples are far from enough to train a robust deep network, multiple works designed single-stream shallow CNNs for MER \cite{gan2018bi,mayya2016combining,gupta2021merastc}.  Belaiche~\etal~\cite{belaiche2020cost} achieved a shallow network through deleting multiple convolutional layers of the deep network Resnet.  Zhao~\etal~ \cite{zhao2020compound} proposed a 6-layer CNN in which the input is followed by an $1\times1$ convolutional layer to increase the non-linear representation.

Besides designing shallow networks, many studies \cite{li2018can,liu2020sma,le2020dynamic} fine-tuned deep networks pre-trained on large face datasets to avoid the overfitting problem. Li~\etal~\cite{li2018can} firstly adopted the 16-layer VGG-FACE model pre-trained on VGG-FACE dataset \cite{Parkhi15} for MER. Following \cite{li2018can}, the MER with  Resnet50, SEnet50 and VGG19 pre-trained on Imagenet was explored in \cite{le2020dynamic}. The results illustrate that VGG surpasses other architectures regarding the MER topic and is good at distinguishing the complex hidden information in data.

%Moreover,  Liu~\etal~\cite{liu2020sma} utilized the backbone of Face Attention Network \cite{wang2017face} pre-trained on WiderFace dataset \cite{yang2016wider} and further added a spatio-temporal movement-attending module to better capture the subtle ME movements. 

%Along with motion magnification and dynamic image computation, subtle movements of ME can be effectively distinguished and discerned.

All of above works are based on 2D CNN with image input, while several works employed single 3D CNN to directly extract the spatial and temporal features from ME sequences. GAM \cite{wang2020eulerian}, MERANet \cite{gajjala2020meranet} and CBAMNet \cite{chen2020spatiotemporal} combined attention modules to 3D CNN to enhance the representation in spatial and temporal dimensions.

%Generally, for inputs, there are two kinds of multiple views: 1) different regions of a face; 2) different modalities of a face, \eg~ frames and optical flow. For structure

\subsubsection{Multi-stream network} 
Single stream is a basic model structure and only extracts features from the single view of MEs. However, MEs have subtle movements and limited samples, the single view is not able to provide sufficient information. 
As we discussed in Section \ref{sec:inputs}, the various inputs from different views is able to effectively explore spatial and temporal information. Thus, the  multi-stream network is adopted in MER to learn features from multiple inputs. The multi-stream structure allows 
the network extracting multi-view features through multi-path networks, as shown in Fig.~\ref{fig:classical} (g). 
In general, multi-stream networks can be classified to networks with the same blocks, different blocks and handcrafted features.

%The learned multi-view features are usually concatenated together to construct a stronger representation for classification. 

%Yan~\etal~\cite{yan2020micro} (Enriched Two Stream 3D Convolutional Network) employed  dual-stream 3D CNN to process the optical flow maps and original frames directly to extract compact temporal information.  

\emph{Multi-stream networks with the same blocks. }
The Optical Flow Features from Apex frame Network (OFF-ApexNet) \cite{gan2019off}  and  Dual-stream shallow network (DSSN) \cite{khor2019dual} built the dual-stream CNN for MER based on optical flow extracted from onset and apex. 
Furthermore, Liong~\etal~\cite{liong2020evaluation} extended OFF-ApexNet to multiple streams with various optical flow components as input data. The   multi-stream CNN with optical flow \cite{liu2020multi} and Three-Stream CNN (TSCNN)~\cite{li2019three,song2019recognizing} designed three-stream CNN models for MER with three kinds of inputs (See Fig.~\ref{fig:classical} (c)). Specifically, the former one utilized apex frame, optical flow and the apex frame masked by the optical flow threshold, while the latter approach employed the apex frames, optical flow between onset, apex, and offset frames to investigate the information of the static spatial, dynamic temporal and local information.
 In addition, She~\etal~\cite{she2020micro} proposed a four-stream model considering three RoIs and global regions as each stream to explore the local and global information.
 Besides multi-stream 2D CNNs, 3DFCNN~\cite{li2019micro}, SETFNet~\cite{9103767} and \cite{yan2020micro} applied 3D flow-based CNNs for video-based MER consisting of multiple sub-streams to extract features from  frame sequences and optical flow, or RoIs.

 %Specifically,  3DFCNN  extracts deeply learned features  with the grayscale frame sequences and the horizontal and vertical components of the optical flow. SETFNet took three facial regions as the inputs of three 3DCNN streams and introduced SE block to learn the weights of each feature channel from the concatenated feature of three streams.

\emph{Multi-stream networks with  different blocks.}  For enhancing the ME feature representation, some works \cite{liong2019shallow,verma2020affectivenet,she2020micro,wang2020micro,wu2021tsnn} investigated the combination of different convolutions.  
Liong~\etal~ designed a Shallow Triple Stream Three-dimensional CNN (STSTNet)~\cite{liong2019shallow}  adopting multiple 2D CNN with different kernels.
Instead of utilizing different kernels, AffectiveNet \cite{verma2020affectivenet} constructed a four-path network with four different receptive fields (RF) to obtain multi-scale features 
for better describing subtle MEs. 
On the other hand, Landmark Relations with Graph
Attention Convolutional Network (LR-GACNN)
 \cite{Kumar_2021_CVPR} and MER-GCN \cite{lo2020mer} built two-stream graph networks to explore relationships between landmark points and the local patches, and AUs and sequence, respectively.  Furthermore, \cite{wang2020micro} and \cite{wu2021tsnn} integrated  2D CNN and 3D CNN to extract spatio-temporal information.

%The spatial and temporal features were concatenated to represent the MEs.

\emph{Multi-stream networks with  handcrafted features.}
Since the  subtle facial movements of MEs are highly related to face textures, the handcrafted features for low-level representation also plays an important role in MER. Multiple works \cite{takalkar2020manifold, hu2018multi,pan2020hierarchical} combined deep features and handcrafted features to leverage the low-level and high-level information for robust MER. Specifically, in the works \cite{takalkar2020manifold} and \cite{pan2020hierarchical}, the CNN features on apex frame and LBP-TOP were concatenated to represent MEs.

\subsubsection{Cascaded network}
%MEs flit spontaneously and has varying duration. Effective temporal analysis is important for ME MER.  Recently, RNN \cite{medsker2001recurrent} is proposed to model sequences  and can work with convolutional layers to model spatio-temporal representation.

Cascaded network combines various modules for different tasks sequentially to construct an effective network, as shown in Fig. \ref{fig:classical} (g). Recent FE studies \cite{li2020deep} demonstrate that learning a hierarchy of features gradually filters out the information unrelated to expressions. 

Inspired by the FE studies \cite{li2020deep}, for further exploring the temporal information of MEs, Nistor~\etal~\cite{nistor2020multi}  cascaded CNN and RNN to  extract features from individual frames of the sequence and capture the facial evolution during the sequence, respectively.  Furthermore, Bai~\etal~\cite{bai2020investigating} and Zhi~\etal~\cite{zhi2019facial} combined CNN with LSTMs in series to deal with ME samples with various duration directly, as shown in Fig.~\ref{fig:classical} (b). Besides, in order to explore the AU semantics in MEs, Xie~\etal~proposed an AU-assisted Graph Attention Convolutional Network (AU-GACN) \cite{xie2020assisted} cascading 3D CNN and GCN to infer MEs based on AU features (see Fig.~\ref{fig:classical} (f)).

%Instead, the Long Short-Term Memory (LSTM)  can resolve vanishing gradient problem and is well-suited to  process the time series with unknown duration

%Considering the redundant information in MEs, instead of employing all of the ME frames, the Adaptive Key-frame Mining Network (AKMNet) \cite{peng2020recognizing} spotted keyframes through attentions and utilized bi-GRU \cite{chung2014empirical} with fewer trainable parameters to compute the essential temporal context information across all keyframes.

%\subsubsection{Multi-stream cascaded network}

In addition, multiple MER works combined  multi-stream and cascaded structure to further explore the multi-view series information. 
VGGFace2+LSTM \cite{bai2020investigating},  Temporal Facial Micro-Variation Network (TFMVN) \cite{zhang2020micro} and MER with Ternary Attentions (MERTA)  \cite{yang2019merta}  developed three stream VGGNets followed by LSTMs to extract multi-view  spatio-temporal features. 
Different from above works,  Khor~\etal~ \cite{khor2018enriched} proposed an Enriched Long-term Recurrent Convolutional Network (ELRCN)  adding one VGG+LSTM path with channel-wise stacking inputs for spatial enrichment. Besides, AT-Net \cite{peng2019novel} and SHCFNet \cite{huang2020shcfnet} extracted spatial and temporal features by CNN and LSTM from the apex frame and optical-flow in parallel and concatenated them together to represent MEs. 

%(optical flow image, optical strain image, and gray-scale raw image)

\subsubsection{Multi-task network}

Most existing works for MER focus on learning features that are sensitive to expressions. However, MEs in the real world are intertwined with various factors, such as subject identity and AUs. The approaches aiming at a single MER task are incapable of making full use of the information on face.  To address this issue, several multi-task learning-based MER approaches have been subsequently developed for better MER \cite{li2019facial,nie2021geme}. Firstly, Li~\etal~\cite{li2019facial} developed a multi-task network combining facial landmarks detection and optical flow extraction to refine the optical flow features for MER with SVM. Following \cite{li2019facial}, several end-to-end deep multi-task networks leveraging different side tasks were proposed. 
 GEnder-based ME recognition (GEME) \cite{nie2021geme} designed a dual-stream multi-task network incorporating gender detection task with MER (see Fig~\ref{fig:classical} (d)), while feature refinement \cite{zhou2021feature} and MER-auGCN \cite{zhou2020objective}  simultaneously detected AUs and MEs and further aggregated AU representation into ME representation. On the other hand, considering that a common feature representation can be learned from multiple tasks, Hu~\etal~\cite{hu2018multi} formulated MER as a multi-task classification problem in which each category classification can be regarded as one-against-all pairwise classification problem.

In summary, the network architecture can be roughly divided into single-stream, multi-stream, cascaded networks, and multi-task learning, as shown in Fig. \ref{fig:classical} (g). Single stream is the simple basic model architecture. However,  single-stream networks only consider the single view of MEs. To further leverage the ME information, the multi-stream network is proposed to learn features from multiple views for robust MER. Moreover, since learning  a  hierarchy  of  features  can  gradually  filter  out the information unrelated to expressions, the network cascades various modules, such as LSTMs and GCNs, sequentially  to  construct  an  effective  MER network. In the future, more effective modules should be combined in multi-stream and cascaded ways to further boost the MER performance. 

In terms of the tasks, multiple task learning \cite{zhang2021survey} can share knowledge  among tasks, introducing extra information and a low risk of overfitting in each task. Currently, most ME research only studied the contribution of landmarks detection, gender classification, and AU detection. Other tasks, such as face recognition and eye gaze tracking may also introduce useful  knowledge for MER. Exploring and taking advantage of  more  face related-tasks is a practical way to further improve MER performance.

%In general, multiple task learning \cite{zhang2021survey} can share knowledge  among tasks, leading to introduce extra information and low risk of overfitting in each task, shown in Fig~\ref{fig:MTL} (b). Currently, most ME research only studied the contribution of the landmarks detection, gender classification and AU detection. Other tasks, such as the face recognition and eye gaze tracking may also introduce useful  knowledge for MER. Exploring and take advantage of  more  face related-tasks is a practical way to further improve MER performance. 

\vspace{-3ex}
\textcolor{black}{ \subsection{Training strategy}}

As we discussed before, DL-based MER suffers from a lack of adequate data. It is almost impossible to train a reliable deep model from scratch. 
Currently,  there are large-scale FE datasets with labels. Leveraging the FE datasets by special training strategy, such as fine-tuning \cite{song2019recognizing}, knowledge distillation~\cite{sun2020dynamic}, and domain adaptation \cite{xia2020learning}, is a reasonable way to solve the problem of a small amount of data.   The knowledge of a pre-trained model for a related task can be transferred to MER to boost performance. The training strategy of transferring knowledge is shown in Fig~\ref{fig:classical} (g).

% Most MER approaches\cite{song2019recognizing,li2020joint,chen2020spatiotemporal,liu2020sma} based on deep networks are pre-trained on ImageNet, FE and face datasets. 

Fine-tuning ME datasets on pre-trained models is widely used in MER  \cite{song2019recognizing,li2020joint,chen2020spatiotemporal,liu2020sma}.
Patel~\etal~\cite{Patel2017Selective} provided two models pre-trained on ImageNet dataset and  FE datasets, respectively. The feature selection method  was also adopted to improve the model's performance. It was found that features captured from the FE datasets performed better in terms of accuracy, as it is more similar to the ME datasets than object/face datasets. %SSPOS and CK+ imilarly, RezNet10 \cite{peng2018macro} architecture pre-trained on ImageNet as initialization is fine-tuned on macro-expression datasets for domain adaptation. 

% Knowledge distillation 
Besides fine-tuning, another effective  transfer learning strategy is knowledge  distillation. Knowledge  distillation achieves  small and fast networks through leveraging information from pre-trained high-capacity networks.  Sun~\etal~\cite{sun2020dynamic}  utilized Fitnets \cite{adriana2015fitnets} to  guide the shallow network learning for  MER by mimicking  the  intermediate  features  of  the  deep network pre-trained for macro-expression recognition and AU detection, as shown in Fig~\ref{fig:classical} (e). However, the  appearances of MEs and  macro-expressions are different due to the different  intensity of  facial movements. Thus, mimicking the macro-expression representation directly is not reasonable.  Instead, SAAT \cite{zhou2019cross}  transferred attention on the style aggregated MEs generated by CycleGAN \cite{zhu2017unpaired}.

In addition,  domain adaptation methods can obtain domain  invariant representations by embedding domain adaptation in the pipeline of deep learning. In \cite{liu2019neural}, \cite{xia2020learning}, and \cite{xiamicro},  the gap between the MEs and macro-expressions was narrowed down by domain adaption based on adversarial learning strategy.

In general, fine-tuning is most widely used in MER. To further effectively transfer meaningful information from massive FEs, knowledge distillation and domain adaptation are also applied to MER by distilling knowledge and extracting domain invariant representations, respectively. However, the domain adaptation with adversarial learning increases the learning complexity. There are significant differences both spatially and temporally between macro-expressions and MEs, therefore, directly transferring the knowledge is not able to fully leverage the macro-expression information. 
Considering that the facial muscle movements are consistent  between  MEs  and  macro-expressions, the attention and AUs can be further studied for transfer learning in future ME research. Moreover, semi-supervised and unsupervised learning could also be further developed to take advantage of unlabeled facial images.

%In fact, LSTM is more commonly used than original RNN and GRU. LSTM also is a variant RNN, and some works[1\cite{khor2018enriched} 3 \cite{zhang2020micro} 4 \cite{yang2019merta} 6\cite{bai2020investigating} 7 \cite{choi2020facial} 8\cite{huang2020shcfnet} 9\cite{zhi2019facial}] connect it with CNN as follows:

%Graph consists of nodes and edges. The node represents the entities in the graph and the edge represents the relations between two entities. The edge set is usually represented as an adjacency matrix A, where each element  degree of relation between the node Ni and Nj . 

%\subsubsection{metric learning} \subsubsection{Margin} \subsubsection{Imbalanced} 
\subsection{Loss functions} 

% In most cases, the final layer of deep network is usually a loss layer which specifies how to penalize the deviation between the predicted and true labels during training. An effective loss function can improve the discriminative ability of the learned representations. During the learning process, the intra-class variations should be minimized and inter-class differences should be maximized. 

%The most commonly used loss function for multi-class classification is the Softmax cross-entropy loss \cite{kline2005revisiting}. The Softmax cross-entropy loss consists of Softmax function and cross-entropy loss. The Softmax function first converts logits into probabilities and then the cross-entropy between the estimated class probabilities and the ground-truth distribution is minimized.  

Different from traditional methods, where the feature extraction and classification are independent, deep networks can perform end-to-end classification through loss functions by penalizing the deviation between the predicted and true labels during training. Most MER works directly applied the commonly used softmax cross-entropy loss \cite{kline2005revisiting}.  The softmax loss is typically good at correctly classifying known categories. However, in practical classification tasks, the unknown samples need to be classified. Therefore, in order to obtain better-generalized ability, the inter-class difference and 
intra-class variation should be further optimized and reduced, respectively, especially for subtle and limited MEs. The metric learning techniques, such as contrastive loss \cite{hadsell2006dimensionality} and triplet loss \cite{schroff2015facenet},  was developed to ensure intra-class compactness and inter-class separability through measuring the relative distances between inputs. Xia~\etal~\cite{xia2020learning} adopted    an adversarial learning approach and triplet loss with inequality regularization to converge the output of MicroNet efficiently.  However, metric learning loss usually requires effective sample mining strategies for robust recognition performance. Metric learning alone is not enough for learning a discriminative metric space for MEs. Intensive experiments demonstrate that importing a large margin  on softmax loss  can increase the inter-class difference.  Lalitha~\etal~\cite{lalitha2020micro} and Li~\etal~\cite{li2020joint} combined softmax cross-entropy loss and center loss \cite{wen2016discriminative} to increase the compactness of intra-class variations  and separable inter-class differences through penalizing the distance between deep features and their corresponding class centers.

Some special MEs are difficult to trigger, thus leading to data imbalance.
Multiple MER works \cite{lai2020real,nie2021geme,xia2019spatiotemporal,zhou2020objective} utilized the Focal loss to overcome the imbalance challenge by adding a factor to put more focus on misclassified  and hard samples which are difficult to recognize. Moreover, MER-auGCN  \cite{zhou2020objective} designed an adaptive factor with the Focal loss to balance the proportion of the negative and positive samples in a given training batch. 

In summary, MER suffers from  high intra-class variation, low inter-class differences, and imbalanced distribution because of the low intensity and spontaneous characteristics of MEs. Currently, most MER approaches are based on the basic softmax cross-entropy loss, but others utilized the triplet loss, center loss, or focal loss to encourage inter-class separability, intra-class compactness, and balanced learning. 
In the future, exploring more effective loss functions  to learn discriminative representation for MEs can be a promising research direction.  Considering the low intensity of facial movements leading to low inter-class differences, 
 better metric space  and larger margin loss for MER should be further studied. Recently, various methods \cite{li2020overcoming} have been proposed for the classification of imbalanced long-tail distribution data. ME research can leverage the ideas for long-tail data to improve the MER performance.

%\begin{figure}
%  \centering
%  \centerline{\includegraphics[width=8.5cm]{flowImg.pdf}}
%\includegraphics[width=\columnwidth]{image/MULTI-TASK2.jpg}
%\caption{(a)  A dual-stream multi-task network incorporating gender detection  designed by GEME \cite{nie2021geme}. (b) Multi-task learning \cite{zhang2021survey}}
%\label{fig:MTL}
%\end{figure}

%Especially for MEs, the subtle facial movements are not able to contribute enough knowledge for robust MER. 

 %In this paper,considering  the  correlation  and  active  facial  regions  are consistent  between  micro-  and  macro-expressions,  ASP  isproposed  to  transfer  the  informative  activation  correlation for  better  supervising  the  training  for  micro-expression  AUdetection  under  domain  shift  without  learning  additionalparameters.

%\begin{figure}
%  \centering
%  \centerline{\includegraphics[width=8.5cm]{flowImg.pdf}}
%\includegraphics[width=\columnwidth]{image/Transfer2.jpg}
%\caption{(a) Pre-training and fine-tuning \cite{Chongyang2020micro} (b) MTMNet achieved by domain adaption based on adversarial learning \cite{xia2020learning} (c) DIKD with knowledge distillation~\cite{sun2020dynamic}  (d) SAAT  based attention transfer and adversarial learning \cite{zhou2019cross} (e) Transfer learning concept}
%\label{fig:Transfer}
%\end{figure}

\subsection{Discussion}
MEs are involuntary, subtle, and brief facial movements.
How to extract high-level discriminative representations on limited subtle ME samples is the main challenge for robust MER with DL. In order to extract discriminative ME representation, various blocks have been designed based on exploring the special characteristics of MEs with less parameters, such as the attention module and capsule module.  In the future, more effective blocks, such as attention, GCN and transformer, should be further developed for MER performance improvement.  On the other hand, considering the limited ME samples, more efficient blocks should be studied to learn discriminative ME features with less parameters for avoiding overfitting on small-scale ME datasets.

In terms of the network architecture, compared with  basic single stream networks,  multi-stream networks can extract features from multi-view inputs to provide more information for MER. On the other hand, the cascaded network combines various modules for different tasks sequentially to construct an effective network and gradually filter out the information unrelated to MEs. Considering the strengths of multi-stream and cascaded networks,  multi-stream cascaded networks have been developed to boost the MER performance further. In the future, exploring multi-stream cascaded networks combined  with various efficient blocks is a promising research direction for MER. In addition, the multi-task learning framework achieves robust MER through leveraging information from related tasks.  
Multi-task learning is able to make  use of more available  information on the face. Current MER explored gender classification, landmark detection, and AU detection to take advantage of existing information as much as possible. In the future, more relevant tasks, such as identity classification and age estimation, could be studied.

Fine-tuning is widely used in MER. Recent research~\cite{sun2020dynamic} illustrated that borrowing information from large FE datasets through knowledge distillation and domain adaptation can achieve promising  performance.  For future ME research, how to effectively leverage massive face images will be a focus. Besides, the semi-supervised learning \cite{van2020survey} and unsupervised learning \cite{wang2018unsupervised} could be  promising research directions.

For the losses, most DL-based MER employs the basic softmax cross-entropy loss. Several works utilized the metric learning loss and margin loss to increase the compactness of intra-class variations and separable inter-class differences. Furthermore, since the ME datasets are imbalanced, multiple works aimed to boost MER performance through Focal loss. However, current MER methods just employed the losses designed for common tasks, such as image classification and face recognition. MER is a special task due to the ME characteristics (low intensity and imbalanced small-scale ME datasets),  effective losses aimed for MER should be explored in the future.

%Taking account of the successful face analysis with improved losses and for further improving the MER performance, the loss can separate the subtle and imbalanced MEs can be considered.

%In the future, on the basis of current structures, the graph learning and transfer learning can be combined to the  multi-stream and cascaded networks 

\section{Experiments}
\label{sec:experiments}

\subsection{Evaluation matrix}
The common evaluation metrics for MER are 
accuracy and F1-score. In general, the accuracy metric measures the ratio of correct predictions over the total evaluated samples. However, the accuracy is susceptible to bias data. F1-score solves the bias problem by considering the total True Positives (TP), False Positives (FP) and False Negatives (FN) to reveal the true classification performance.

For the composited dataset which combines multiple datasets leading to severe data imbalance, Unweighted F1-score (UF1) and Unweighted Average Recall (UAR) are utilized to measure the performance of various methods.  UF1 is also known as macro-averaged F1-score which is determined by averaging the per-class F1-scores. UF1 provides equal emphasis on rare classes in imbalanced multi-class settings. UAR is defined as the average accuracy of each class divided by the number of classes without consideration of  samples per class. UAR can reduce the bias caused by class imbalance and is known as balanced accuracy.

\begin{table*}
\renewcommand{\arraystretch}{1.3}
\centering 
\small
\caption{MER on SMIC, CASME, CASME II, SAMM, and CMED datasets. }

\scalebox{0.70}{ 
\begin{tabular}{|c|c|c|c|c|c|c|c|c|c|c|c|c|}
 \bottomrule[1.5pt]
Dataset&Method&Year&Pre-p.&Input& Network architecture & Block &Pre-train & Protocol&Cate.& F1&ACC (\%)\\
 \bottomrule[1.5pt]
\multirow{8}{*}{SMIC}%&3D-CNN+LSTM \cite{zhi2019facial} &2019& - &Sequence &3DCNN+LSTM&-&-&LOSO&3&- &56.6\\\cline{2-12}

%&OFF-ApexNet \cite{gan2019off} &2019 &-& OF &2S-CNN&-& -&LOSO&3&0.6709& 67.68\\\cline{2-12}

&TSCNN~\cite{song2019recognizing}& 2019&E, R & OF+Apex &3S-CNN&-& FER2013 \cite{goodfellow2013challenges}&LOSO&3&0.7236&72.74\\\cline{2-12}

%&LEARNet~\cite{verma2019learnet} &2019&-&DI &CNN&Hybrid& -&LOSO&3&-& 81.60
%\\\cline{2-12}

%&3D-FCNN~\cite{li2019micro} &2019&T&OF &3S-CNN&-& -&LOSO&3&-&55.49\\\cline{2-12}

%&STRCN-G \cite{xia2019spatiotemporal} &2019&E & OF &CNN&RCN& -&LOSO&3&0.695 &72.3\\\cline{2-12}

%&STSTNet ~\cite{liong2019shallow} &2019 &E & OF &3S-3DCNN&-& -&LOSO&3& 0.6801 &70.13\\\cline{2-12}

%&LGCcon\cite{li2020joint} &2020 &E, R  & Apex &CNN&Attention& VGG-FACE \cite{Parkhi15}&LOSO&3&0.62 & 63.41\\\cline{2-12}

%&CBAMNet \cite{chen2020spatiotemporal} &2020&E, T & Sequence &3DCNN&Attention&Oulu-CASIA  \cite{zhao2011facial}&10-fold&3&  -& 54.84\\\cline{2-12}

%&DSSN \cite{khor2019dual} &2019&-& OF &2S-CNN&-& -&LOSO&3&0.6462& 63.41\\\cline{2-12}

&DIKD~\cite{sun2020dynamic} &2020&-&Apex &CNN+KD+SVM&-& -&LOSO&3&0.71&76.06
\\\cline{2-12}

%&SMA-STN~\cite{liu2020sma} &2020 &-&\textcolor{black}{Snippet}  &CNN&-&WIDER FACE\cite{yang2016wider}&LOSO&3& 0.7683& 77.44\\\cline{2-12}

%&EM-C3D+GAM \cite{wang2020eulerian} &2020&E &  Sequence &3DCNN&Attention& -&LOSO&4& & 69.76\\\cline{2-12}

%&SETFNet \cite{9103767} &2020&R &  Sequence &3S-3DCNN&SE& -&5-Fold&5&-& 70.25\\\cline{2-12}

&MTMNet \cite{xia2020learning} &2020 &- & Onset-Apex &\textcolor{black}{2S-CNN}+DA+GAN&RES& \makecell[c]{CK+ \cite{lucey2010extended},MMI \cite{pantic2005web}, \\ Oulu-CASIA \cite{zhao2011facial}}&LOSO&3&0.744&76.0\\\cline{2-12}

%&GEME  \cite{nie2021geme} &2021&- & DI &\textcolor{black}{2S-CNN}+ML  &RES& -&LOSO&3&0.6158&64.63\\\cline{2-12}

&MiMaNet \cite{xiamicro}&2021&T& Apex+sequence &\textcolor{black}{2S-CNN}+DA & RES& CK+ \cite{lucey2010extended},MMI\cite{pantic2005web}&LOSO&3&0.778&78.6\\\cline{2-12}

& DSTAN\cite{wang2021micro}&2021&T& OF+sequence &\textcolor{black}{2S-CNN}+LSTM+SVM & Attention& -&LOSO&3&0.78&77\\\cline{2-12}

& AMAN\cite{wei2022novel}&2022&E,T& sequence &\textcolor{black}{CNN} & Attention& FER2013 \cite{goodfellow2013challenges}&LOSO&3&0.77&79.87

\\ \bottomrule[1.5pt]

%&KFC \cite{su2021key} &2021&-& OF &\textcolor{black}{2S-CNN} & Attention& -&LOSO&3&0.6638&65.85

\multirow{5}{*}{CASME} & TSCNN~\cite{song2019recognizing} &2019& E,R & OF+Apex &3S-CNN&-& FER2013 \cite{goodfellow2013challenges}&LOSO&4&0.7270 & 73.88
\\\cline{2-12}

%&LEARNet~\cite{verma2019learnet} &2019&-&DI &CNN&Hyfeat& -&LOSO&8&-& 80.62\\\cline{2-12}

%SCNN~\cite{miao2019recognizing} &2019&-&OF &CNN&-& -&LOSO&5&0.6142& 63.33
%\\\hline

%&3D-FCNN~\cite{li2019micro} &2019&T&OF &3S-3DCNN&-& -&LOSO&5&-&54.44\\\cline{2-12}

%&LGCcon\cite{li2020joint} &2020&E, R & Apex &CNN&Attention& VGG-FACE \cite{Parkhi15}&LOSO&4&0.60&60.82 \\\cline{2-12}

&DIKD~\cite{sun2020dynamic} &2020&-&Apex &CNN+KD+SVM&RES& -&LOSO&4&0.77&81.80
\\\cline{2-12}

%&OrigiNet~\cite{verma2020non} &2020 & -&\textcolor{black}{AI}  &CNN&Hyfeat&-&LOSO&4& -& 66.09\\\cline{2-12}

%MANet~\cite{she2020micro} Apex
%\\\hline

&AffectiveNet \cite{verma2020affectivenet}
&2020&E & DI &4S-CNN&MFL& -&LOSO&4&-&72.64\\\cline{2-12}

& DSTAN\cite{wang2021micro}&2021&T& OF+sequence &\textcolor{black}{2S-CNN}+LSTM+SVM & Attention& -&LOSO&4&0.75&78\\

 \bottomrule[1.5pt]

\multirow{11}{*}{CASME II} %&\textcolor{black}{ELRCN} \cite{khor2018enriched}&2018 &T& OF &4S-CNN+LSTM&-&VGG-Face \cite{Parkhi15}&LOSO&5&0.5&52.44\\\cline{2-12}

&OFF-ApexNet \cite{gan2019off} &2019 
&-& OF &2S-CNN&-& -&LOSO&3&0.8697& 88.28\\\cline{2-12}

&TSCNN~\cite{song2019recognizing} &2019&E, R & OF+Apex &3S-CNN&-& FER2013 \cite{goodfellow2013challenges}&LOSO&5&0.807&80.97\\\cline{2-12}

%&LEARNet~\cite{verma2019learnet} &2019&-&DI &CNN&Hyfeat& -&LOSO&7&-& 76.57\\\cline{2-12}

%SCNN~\cite{miao2019recognizing} &2019&-&OF &CNN&-& -&LOSO&5&0.6981& 71.14
%\\\hline

%&3D-FCNN~\cite{li2019micro} &2019&T&OF &3S-CNN&-& -&LOSO&5&-&59.11\\\cline{2-12}

%&STRCN-G \cite{xia2019spatiotemporal}& 2019&E & OF &CNN&RCN& -&LOSO&3&0.747 &80.3\\\cline{2-12}

&STSTNet~\cite{liong2019shallow} &2019
&E & OF &3S-3DCNN&-& -&LOSO&3& 0.8382& 86.86\\\cline{2-12}

%&DSSN \cite{khor2019dual} &2019
%&-& OF &2S-CNN&-& -&LOSO&5&0.7297 & 71.19\\\cline{2-12}

%&3D-CNN+LSTM \cite{zhi2019facial} &2019&-&Sequence &3DCNN+LSTM&-&-&LOSO&5&- &62.5\\\cline{2-12}

&Graph-TCN~\cite{lei2020novel} & 2020 &L, R &Apex &TCN+GCN&Graph&- &LOSO&5&0.7246 & 73.98\\\cline{2-12}

%&AU-GACN \cite{xie2020assisted} &2020 & - & Sequence & 3DCNN+GCN& Graph&- &{LOSO}&3&0.355&71.2\\\cline{2-12}

%&CBAMNet \cite{chen2020spatiotemporal}&2020 &E, T & Sequence &3DCNN&Attention & Oulu-CASIA \cite{zhao2011facial}&10-fold&3&  -& 69.92\\\cline{2-12}

%&LGCcon\cite{li2020joint} &2020 &E, R  & Apex &CNN&Attention& VGG-FACE \cite{Parkhi15}&LOSO&5&0.64 & 65.02\\\cline{2-12}

%&CNNCapsNet~\cite{liu2020offset} &2020 &- & OF &5S-CNN&Capsule& -&LOSO&5&0.6349 & 64.63\\\cline{2-12}

%&DIKD~\cite{sun2020dynamic} &2020&-&Apex &CNN+KD+SVM&-& -&LOSO&4&0.67&72.61\\\cline{2-12}

%&OrigiNet~\cite{verma2020non} &2020&-&\textcolor{black}{AI}  &CNN&Hyfeat&-&LOSO&4& -& 62.09\\\cline{2-12}

&SMA-STN~\cite{liu2020sma} &2020 
&-&\textcolor{black}{Snippet}  &CNN&Attention&WIDER FACE\cite{yang2016wider}&LOSO&5& 0.7946& 82.59\\\cline{2-12}

%&EM-C3D+GAM \cite{wang2020eulerian} &2020&E &  Sequence &3DCNN&Attention& -&LOSO&4& & 69.76\\\cline{2-12}

%&SETFNet \cite{9103767}& 2020 &R &  Sequence &3S-3DCNN&SE& -&5-Fold&5&-& 66.28\\\cline{2-12}

%&MSCNN \cite{liu2020multi} &2020&E & OF+Apex &3S-CNN&RES& -&LOSO&5&0.5325& 56.50\\\cline{2-12}

%&AffectiveNet \cite{verma2020affectivenet}&2020 &E & DI &4S-CNN&MFL& -&LOSO&4&-&68.74\\\cline{2-12}

%ELRCN \cite{khor2018enriched} & & Sequence &CNN+LSTM&VGG-Face&LOSO&5&0.5&52.44\\\hline

%ATNet& & OF+Apex &CNN+LSTM&FE database&LOSO&3&0.798&83.4\\\hline

&GEME  \cite{nie2021geme} &2021&- & DI &\textcolor{black}{2S-CNN}+ML&RES& -&LOSO&5&0.7354&75.20\\\cline{2-12}

&MiMaNet \cite{xiamicro}&2021&T& Apex+sequence &\textcolor{black}{2S-CNN}+DA & RES& CK+ \cite{lucey2010extended},MMI\cite{pantic2005web}&LOSO&5&0.759&79.9\\\cline{2-12}

&LR-GACNN \cite{Kumar_2021_CVPR}&2021&E& OF+Landmark &\textcolor{black}{2S-GACNN} & Graph& -&LOSO&5&0.7090&81.30\\\cline{2-12}

%&GRAPH-AU \cite{Lei_2021_CVPR} &2021&L& Apex &\textcolor{black}{2S-CNN+GCN} & \makecell[c]{Graph, \\ Transformer}& -&LOSO&5&0.7047&74.27\\\cline{2-12}

& DSTAN\cite{wang2021micro}&2021&T& OF+sequence &\textcolor{black}{2S-CNN}+LSTM+SVM & Attention& -&LOSO&5&0.73&75\\\cline{2-12}

& AMAN\cite{wei2022novel}&2022&E,T& sequence &\textcolor{black}{CNN} & Attention& FER2013 \cite{goodfellow2013challenges}&LOSO&5&0.71&75.40\\\bottomrule[1.5pt]

%&KFC \cite{su2021key} &2021&-& OF &\textcolor{black}{2S-CNN} & Attention& -&LOSO&5&0.7375&72.76\\ \bottomrule[1.5pt]

\hline
%%%%%%%%%%%%%%%%%%%%%%%%%%%%%%%%%%%%%%%%%%%%%%

\multirow{8}{*}{SAMM}%&OFF-ApexNet \cite{gan2019off} &2019 & & OF &2S-CNN&-&-&LOSO&3&0.5423& 68.18\\\cline{2-12}

%AU-GACN & - & Sequence & 3DCNN+Graph(AU)  &\multirow{-} &\multirow{LOSO}&7&0.357&52.3\\
%&&&&&&3&0.433&70.2\\
%&&&&&\multirow{LOVO}&7&0.228&42.6\\
%&&&&&&3&0.454&72.1\\\hline

%&STSTNet ~\cite{liong2019shallow} &2019 &E & OF &3S-3DCNN&-& -&LOSO&3& 0.6588& 68.10\\\cline{2-12}

%&Graph-TCN~\cite{lei2020novel} &2020& L,R &Apex &TCN+GCN&Graph&- &{LOSO} &5& 0.6985 &75.00\\\cline{2-12}

%&AU-GACN~\cite{xie2020assisted}&  2020& - & Sequence & 3DCNN+GCN& Graph &- &{LOSO}&3&0.433&70.2\\\cline{2-12}

%&LGCcon\cite{li2020joint}& 2020 &E, R  & Apex &CNN&Attention& VGG-FACE \cite{Parkhi15}&LOSO&5&0.34 & 40.90\\\cline{2-12}

%&STRCN-G \cite{xia2019spatiotemporal} &2019&E & OF &CNN&RCN& -&LOSO&3&0.741 &78.6\\\cline{2-12}

%&TSCNN~\cite{song2019recognizing}& 2019&E, R & OF+Apex &3S-CNN&-& FER2013 \cite{goodfellow2013challenges}&LOSO&5&0.6942&71.76\\\cline{2-12}

&DIKD~\cite{sun2020dynamic} &2020&-&Apex &CNN+KD+SVM&-& -&LOSO&4&0.83&86.74
\\\cline{2-12}

%&OrigiNet~\cite{verma2020non} &2020&-&AI &CNN&Hyfeat&-&LOSO&4& -&34.89\\\cline{2-12}

&SMA-STN~\cite{liu2020sma} &2020 
&-&\textcolor{black}{Snippet}  &CNN&-&WIDER FACE\cite{yang2016wider}&LOSO&5& 0.7033& 77.20\\\cline{2-12}

%&DSSN \cite{khor2019dual} &2019&-& OF &2S-CNN&-& -&LOSO&5&0.4644 & 57.35\\\cline{2-12}

%&MSCNN \cite{liu2020multi} &2020&E & OF+Apex &3S-CNN& RES& -&LOSO&5&0.3569& 43.04\\\cline{2-12}

%ATNet& & OF+Apex &CNN+LSTM&facial expression database&LOSO&3&0.496&70.1\\\hline

&MTMNet \cite{xia2020learning} &2020 &- & Onset-Apex &\textcolor{black}{2S-CNN}+GAN+DA&RES& 
\makecell[c]{CK+ \cite{lucey2010extended},MMI \cite{pantic2005web}, \\ Oulu-CASIA \cite{zhao2011facial}}&LOSO&5&0.736&74.1\\\cline{2-12}

%&GEME  \cite{nie2021geme} &2021&- & DI &\textcolor{black}{2S-CNN+ML}  &RES & -&LOSO&5&0.4538&55.88\\\cline{2-12}

&MiMaNet \cite{xiamicro}&2021&T& Apex+sequence &\textcolor{black}{2S-CNN}+DA & RES& CK+ \cite{lucey2010extended},MMI\cite{pantic2005web}&LOSO&5&0.764&76.7\\\cline{2-12}

&LR-GACNN \cite{Kumar_2021_CVPR}&2021&E& OF+Landmark &\textcolor{black}{2S-GACNN} & -& -&LOSO&5&0.8279&88.24\\\cline{2-12}

&GRAPH-AU \cite{Lei_2021_CVPR} &2021&L& Apex &\textcolor{black}{2S-CNN+GCN} & \makecell[c]{Graph, \\ Transformer}& -&LOSO&5&0.7045&74.26\\\cline{2-12}

& AMAN\cite{wei2022novel}&2022&E,T& sequence &\textcolor{black}{CNN} & Attention& FER2013 \cite{goodfellow2013challenges}&LOSO&5&0.67&68.85\\

%&KFC \cite{su2021key} &2021&-& OF &\textcolor{black}{2S-CNN} & Attention& -&LOSO&5&0.5709&63.24\\ 

\bottomrule[1.5pt]

CMED& Shallow CNN\cite{zhao2020compound} &2020&E& OF &\textcolor{black}{CNN} & -& -&LOSO&7&0.6353&66.06\\\bottomrule[1.5pt]

%%%%%%%%%%%%%%%%%%%%%%%%%%%%%%%%%%%%
\end{tabular}

}

{\raggedright $^1$   Pre-p.: Pre-processing;  E:EVM; R: RoI; T: Temporal normalization ; L: Learning-based magnification. \par 
$^2$ OF: Optical flow; DI: Dynamic image.\par
$^3$ nS-CNN: n-stream CNN; ML: Multi-task learning; DA: Domain adaption; KD: Knowledge distillation.\par
$^4$   Cate: Category; F1: F1-score; ACC: Accuracy; RES: Residual block.\par
 }
\vspace{-2ex}
 \label{tab:result_CASME}
\end{table*}

\subsection{Model evaluation protocols}
Cross-validation is the widely utilized protocol for evaluating the MER performance. In cross-validation, the dataset is splitted into multiple folds and the training and testing were evaluated on different folds. It regards a fair verification and prevents overfitting on the small-scale ME datasets.  In the MER field, cross-validation includes leave-one-subject-out (LOSO), leave-one-video-out (LOVO), and K-Fold cross-validations.

In LOSO, every subject is taken as a test set in turn and the other subjects as the training data. This kind of subject-independent protocol can avoid subject bias and evaluate the generalization performance of various algorithms.  LOSO is the most popular cross-validation in MER. 

The LOVO takes each sample as the validation unit which enables more training data and alleviates the overfitting to some degree.  However, it is not subject-independent, thus it can not well evaluate the generalization capability. Another problem is that the test number of LOVO is the sample size which may lead to huge time cost, not suitable for deep learning.

For K-fold cross-validation, the original samples are randomly partitioned into k equal-sized parts. Each part is taken as a test set in turn and the rest are the training data. Thus, the number of cross-validation tests is K. In practice,  the evaluation time can be greatly reduced by setting an appropriate K. The typical K values are 5 or 10.

Since the MEs have small-scale datasets, the experiments on MER do not have reliable validation datasets. According to the released codes, some works \cite{liu2019neural} utilized the test datasets as the validation datasets directly and reserved the best epoch results on each fold as the final results. As the data is limited, even only two samples for some subjects, the final MER results will be greatly improved by regarding the test data as the validation data. 
According to \cite{xia2020revealing}, compared to the experiments based on the same epoch on all of the folds, the results can be increased by more than 10$\%$ by testing on the test datasets. 
But, in practice, the test data is unknown and it is not reasonable to reserve the best epoch results on each fold of the test data as the final results.

%\textcolor{red}{In the future, as the validation dataset is missing, the experiments maybe relied on the equal epoch on of the folds according to the change of training loss.}

\begin{table*}
\renewcommand{\arraystretch}{1.3}
\centering 
\small
\caption{MER on the Composite dataset (MECG2019)}

\scalebox{0.80}{ 
\begin{tabular}{|c|c|c|c|c|c|c|c|c|c|c|c|c|c|}
 \bottomrule[1.5pt]
Method&Year&Pre-p.&Input& Network architecture& Block &Pre-train & Protocol&Cate.& UF1&UAR\\ \bottomrule[1.5pt]

%\textcolor{black}{SHCFNet}~\cite{takalkar2017image} &2017& - & OF+Apex&4S-CNN+LSTM&Hyfeat&- &LOSO&3&0.6242&0.6222\\\hline

%OFF-ApexNet \cite{gan2019off} &2019&-& OF &2S-CNN&-& -&LOSO&3& 0.7104 & 0.7460\\\hline

%STSTNet ~\cite{liong2019shallow} &2019&E & OF &3S-3DCNN&-& ImageNet&LOSO&3&0.735&0.760 \\\hline

NMER ~\cite{liu2019neural} &2019&E, R & OF &CNN+DA&-& -&LOSO&3&0.7885&0.7824   \\\hline

%CapsuleNet~\cite{van2019capsulenet} &2019& -& Apex &CNN&Capsule&ImageNet&LOSO&3& 0.6520& 0.6506 \\\hline

Dual-Inception~\cite{zhou2019dual} &2019&- &OF &2S-CNN&Inception &-&LOSO&3&0.7322& 0.7278\\\hline

ICE-GAN~\cite{yu2020ice} &2020& GAN & Apex &CNN+GAN&Capsule &ImageNet &LOSO&3&0.845&0.841\\\hline

%\cite{sabour2017dynamic}
%RCN-A~\cite{xia2020revealing} &2020& E & OF &CNN&RCN&ImageNet &LOSO&3&0.7190&0.7432\\\hline

%ATNet \cite{peng2019novel} 2019& & OF+Apex &CNN+LSTM RES10&CK+, Oulu-CASIA, Jaffe, MUGFE &LOSO&3&0.553&56.1\\\hline

MTMNet \cite{xia2020learning} &2020 &- & Onset-Apex &\textcolor{black}{2S-CNN}+GAN+DA  &RES& \makecell[c]{CK+ \cite{lucey2010extended},MMI \cite{pantic2005web}, \\ Oulu-CASIA \cite{zhao2011facial}}&LOSO&3&0.864&0.857\\\hline

%GEME   \cite{nie2021geme} &2021&- & DI &\textcolor{black}{2S-CNN}+ML & RES& -&LOSO&3&0.7221&0.7303\\\hline

FR \cite{zhou2021feature} &2021&- & OF &\textcolor{black}{2S-CNN}+ML & Inception & -&LOSO&3&0.7838&0.7832\\\hline
%\cite{szegedy2017inception}

MiMaNet \cite{xiamicro}&2021&T& Apex+sequence &\textcolor{black}{2S-CNN}+DA & RES& CK+ \cite{lucey2010extended},MMI\cite{pantic2005web}&LOSO&3&0.883&0.876\\\hline

GRAPH-AU \cite{Lei_2021_CVPR} &2021&L& Apex &\textcolor{black}{2S-CNN+GCN} & \makecell[c]{Graph, \\ Transformer }& -&LOSO&3&0.7914&0.7933\\\hline
%\cite{parmar2018image}

BDCNN \cite{chen2022block} &2022&L& OF &\textcolor{black}{4S-CNN} & \makecell[c]{-}&
-&LOSO&3&0.8509&0.8500

\\ \bottomrule[1.5pt]

\end{tabular}
}

{\raggedright $^1$   Pre-p.: Pre-processing;  E: EVM; R: RoI; T: Temporal normalization ; L: Learning-based magnification. \par 
$^2$ OF: Optical flow; DI: Dynamic image.\par
$^3$  nS-CNN: n-stream CNN; ML: Multi-task learning; DA: Domain adaption; KD: Knowledge distillation\par
$^4$  Cate.: Category; RES: Residual block.\par
 }
 \label{tab:result_MER2019}
\end{table*}

Tables~\ref{tab:result_CASME} and~\ref{tab:result_MER2019} list the reported performance of representative recent work of DL-based MER on popular ME datasets. As we discussed before, the evaluation protocol is varying and the practical training rule of each paper is ambiguous, we can not directly make a conclusion that which method performs best for MER. But, from the experimental results, the general trends of MER can be found. 

For the input, in general, the combined inputs can provide promising results on all of the datasets \cite{song2019recognizing,wang2021micro,Kumar_2021_CVPR}. This is because the different input modalities can contribute information from different views. On the basis of various input modalities, we can explore useful information on limited ME samples to the greatest extent. Since the combined inputs is a good choice for robust MER, the multi-stream network is recommended to learn effective representations from various inputs \cite{song2019recognizing,wang2021micro,Kumar_2021_CVPR}. In contrast to the combined inputs, the sole sequence performs worse \cite{wei2022novel}, due to the limited information and redundancy.

Besides, from Tables~\ref{tab:result_CASME} and~\ref{tab:result_MER2019}, it can be seen that the learning strategy including fine-tuning \cite{liu2020sma}, domain adaptation \cite{xia2020learning, xiamicro} and knowledge distillation \cite{sun2020dynamic} can achieve state-of-the-art results on both the individual datasets and the composite dataset.  This could be explained that the limited ME sample is the main challenge for MER and leveraging other related data sources is a reasonable and effective solution. In the future, domain adaption and knowledge distillation should be further researched to boost MER performance.

In some latest works \cite{Lei_2021_CVPR,Kumar_2021_CVPR,lei2020novel}, the GCN becomes a mainstream choice for MER and shows promising performance. Currently, the spatio-temporal graph representation combined with GCNs obtains more attention in MER studies. The possible reason is that the landmark and AU information are helpful and effective for locating and representing the facial muscle movements. However, the small-sample ME datasets limit the ability of graph representation. The combination of transfer learning and graph should be a promising direction for future ME studies. 

%A problem in graph-based MER is the lack of largescale in-the-wild databases. 

%\begin{table}[!t]
%\renewcommand{\arraystretch}{1.3}
% if using array.sty, it might be a good idea to tweak the value of
% \extrarowheight as needed to properly center the text within the cells
%\caption{An Example of a Table}
%\label{table_example}
%\centering
% Some packages, such as MDW tools, offer better commands for making tables
% than the plain LaTeX2e tabular which is used here.
%\begin{tabular}{|c||c|}
%\hline
%One & Two\\
%\hline
%Three & Four\\
%\hline
%\end{tabular}
%\end{table}

\section{Challenges AND FUTURE DIRECTIONS}
\label{sec:Challenge}
\textcolor{black}{MER has a wide range of potential applications in various fields, such as psychological disorders, education, business negotiation, and security control. More specific descriptions about applications of MER are introduced in APPENDIX A. Although MER could facilitate society in various fields, there are many challenges. In this section, we discuss the challenges and future directions of MER. }

%\textcolor{black}{Psychological disorder: Through detecting and analyzing FEs and MEs, it is possible to identify the true emotions of patients with psychological disorders. Experts can help the patients with early diagnosis and intervention with the assistance of ME and FE analysis. }

%\textcolor{black}{Online education: Integrating MER into the online education system can help tutors to be capable of adjusting the teaching styles regarding the engagement, interests, and distraction, et al., of students in the class and improve teaching effectiveness.}

%\textcolor{black}{ Business negotiation: MER can provide important clues about the real feeling. For example, business negotiators can use glimpses of happiness to determine when they have proposed a suitable price. Based on the clues, the negotiation process can be well mastered, and the goal could be achieved. }

%\textcolor{black}{ Security control: MER can help security personnel identify suspicious behavior and safeguard social security.}

%\textcolor{black}{More specific descriptions about applications of MER are introduced in APPENDIX A. Although MER could facilitate society in various fields, there are many challenges. In this section, we discuss the challenges and future directions of MER. }

\subsection{Dealing with small-scale dataset}
\textcolor{black}{
Deep learning is a data-driven method, and successful training requires various large-scale data. Recent studies indicated that annotation bias, emotional contexts, and cultural backgrounds could affect the ME perception \cite{li2022cas}. They may mislead the model training and finally cause misclassification. Unfortunately, existing ME datasets are far from enough for training a robust model. To this end, more diverse ME datasets should be collected. Besides, effective deep-based data augmentation approaches should be further developed for ME analysis to avoid over-fitting.  Semi-supervised and unsupervised learning could also be potential solutions.
} 

In addition, some emotions, such as fear, are challenging to be evoked and collected.  The data imbalance causes the network to be biased towards classes in the majority. Therefore, effective imbalanced losses are needed.

\subsection{3D ME sequence}
Currently,  the main focus of MER is based on the 2D domain because of the data prevalence in the relevant modalities including images and videos. Although significant progress has been made for MER in recent years, most existing MER algorithms based on 2D facial images and sequences can not solve the challenging problems of illumination and pose variations in real-world applications. Recent research about FE analysis illustrates that the above issues can be addressed through 3D facial data \cite{sandbach2012static}. Inherent characteristics of 3D face make facial recognition robust to lighting and pose variations. Moreover, 
3D geometry information may include important features for FER and provide more data for better training. Thanks to the benefits of 3D faces and the technological development of 3D scanning,
MER based on 3D sequence could be a promising research direction. Special 3D blocks, such as 3D Graph and Transformer, should be studied in 3D MER.

\subsection{AU analysis in MEs}
MEs reveal people's hidden emotions in high-stake situations~\cite{Ekman:1969,Ekman:2009} and have various applications such as clinic diagnosis and national security. However, ME interpretation suffers ambiguities ~\cite{davison2018objective}, \eg, the inner brow raiser may refer to surprise or sad.  The FACS \cite{friesen1978facial} has been verified to be effective for resolving the ambiguity issue. In FACS, action units (AUs) are defined as the basic facial movements, working as the building blocks to formulate multiple FEs~\cite{friesen1978facial}. 
Furthermore, the criteria for AU and FE correspondence is defined in FACS manual. Encoding AUs has been verified to  benefit the MER~\cite{sun2020dynamic, xie2020assisted,chen2021understanding} through embedding AU features. 
In the future, the relationship between AUs and MEs can be further explored to improve MER.

\subsection{\textcolor{black}{Multi-modal MER}}
% With the developments of social media platforms, a huge amount of multi-modal data such as texts, images, audio files, and videos are generated on these platforms simultaneously. 
One of the  MER challenges is that the low-intensity and small-scale ME datasets provide very limited information for robust MER. Recent research demonstrated that utilizing multiple modalities can provide complementary information and enhance classification robustness.  Different emotional expressions can produce different changes in autonomic activity, \eg~fear leads to increased heart rate and decreased skin temperature.  Thus,  the physiological signal can be utilized to  incorporate complementary information for further improving MER. Besides, in recent years, new micro-gesture datasets \cite{chen2019analyze} had been proposed. 
 The micro-gesture is body movements that
are elicited when hidden expressions are triggered in unconstrained situations. The hidden emotional states can be reflected through micro-gestures. How to combine multiple modalities to enhance MER performance is an important future direction. Lightweight multi-stream networks should be developed to learn multi-view ME information effectively and efficiently.

\textcolor{black}{\subsection{The explainanty of MER based on DL }
The neural network is a brain-inspired model developed by neurobiologists and psychologists to test the computational analog of neurons \cite{arrieta2020explainable}. Naturally, it could be a tool to verify the theory in other disciplines, such as psychology, to enhance psychological and human communication study. In addition, the current DL is a “black box” algorithm \cite{arrieta2020explainable} and focuses on learning features and recognizing patterns by updating the weights of networks. The interpretation and understanding of the inside DL process can get experts from cross disciplines involved in the internal state analysis and therefore facilitate building interpretable and reliable deep models. }

\subsection{MEs in realistic situations}
Currently, most existing MER researches focus on classifying the basic MEs collected in controlled environments from the frontal view without any head movements, illumination variations or occlusion. However, it is almost impossible to reproduce such strict conditions in real-world applications. The approaches based on the constrained settings usually do not generalize well to videos recorded in-the-wild environment. Practical and robust algorithms for recognizing MEs in  realistic situations with pose changes and illumination variations should be developed in the future. 

Moreover,  most ME researches assume that there are just MEs in a video clip. However, in real life,  MEs can appear with macro-expressions. Future studies should explore deep-based ME spotting methods to detect and distinguish the micro- and macro-expressions when they occur at the same time.  Analyzing the macro and micro-expressions simultaneously would be helpful to understand people’s intentions and feelings in reality  more accurately.

%Furthermore, in most ME dataset collection, participants were asked to keep neutral faces when watching emotion-stimulating movie clips. The conflict of emotions elicited by the emotional movie clips and the intention to suppress FEs could induce involuntary MEs. In this way,  the collected video clips including MEs are unlikely to have other natural FEs. 

%\textcolor{black}{in Section 7.6 describing how the proposed systems can be eventually employed in applications that pertain to health and education. Please introduce an application section to highlight the potential application scenarios }

%have great application value in the fields of international counter-terrorism, national security, and judicial interrogation. 

\vspace{-2ex}

\textcolor{black}{
\section{Ethical considerations}
As discussed above, MEs can help reveal people’s hidden feelings in high-stake situations and have practical applications in various fields, such as medical treatment and interrogations. MER, like many other computer vision and machine learning tasks,  could be misused, especially when used in surveillance with predatory data collection practices \cite{kosinski2021facial,zuboff2019age,raji2020saving,oviatt2021technology} Therefore, ethical issues should be considered.  
 }

\textcolor{black}{
Privacy and data protection is the primarily and frequently discussed ethical issue in machine learning. For MER, the critical privacy concern is the privacy of personal data. Currently, data protection laws are well established to regulate data privacy, for example, the EU General Data Protection Regulation (GDPR) \cite{voigt2017eu}. The legislation defined rules for the protection of personal data, including international data protection agreements, privacy shields, transfer of participant names, record data, etc. In the research community, consent forms concerning data collection, processing, and sharing need to be signed when collecting ME data. In practical applications, consent forms should also be considered to regulate the usage, as people’s faces are present in the recorded images/videos with sensitive and biometric information that may be misused beyond the intended purpose.  Pilot studies aim to remove sensitive information like identity while preserving facial properties \cite{proencca2021uu}, which could be further explored in MER.}

%the California Consumer Privacy Act (CCPA) \cite{de2018guide} and

\textcolor{black}{
Moreover, questions of reliability in MER systems are further pointed out together with privacy and data protection \cite{stahl2021ethical}. Results of a deep learning-based MER system usually depend on the quality of training data, which are difficult to ascertain because of possible data biases. Transparency of data and models should be aware and well-studied.}

\section{Acknowledgement}
This work was supported by the Academy of Finland for Academy Professor project EmotionAI (grants 336116, 345122), by Ministry of Education and Culture of Finland for AI forum project and Infotech Oulu.

% or



%\ifCLASSOPTIONcompsoc
%\else
%\fi

\ifCLASSOPTIONcaptionsoff
  \newpage
\fi



%\bibliographystyle{IEEEtran}

\bibliographystyle{IEEEtran}
\bibliography{main}

%\begin{thebibliography}{1}


%\end{thebibliography}



%\vfill


\end{document}